\documentclass[12pt]{article}

\usepackage[headheight=15pt]{geometry}
\usepackage{package/tech-report}
\usepackage{import}
\usepackage{array}
\usepackage{amsmath}
\usepackage{amssymb}
\usepackage{booktabs}
\usepackage{multirow}
\usepackage{graphicx}
\usepackage{hyperref}
\usepackage{caption}
\usepackage{enumitem}
\usepackage{xcolor}
\usepackage{caption}
\usepackage{subcaption}
\usepackage{microtype}  % Better line breaking and typography
\usepackage[all]{nowidow}  % Prevent widows and orphans
\usepackage{upgreek,textgreek}
\usepackage{tikz,lipsum,lmodern}
\usepackage[most]{tcolorbox}
\usepackage{listings}
\usepackage{cleveref}
\usepackage[flushleft]{threeparttable}
\usepackage{titletoc}
\usepackage{minted}
\usepackage{algorithm} % For the algorithm environment
\usepackage{algpseudocode} % For algorithmic pseudocode
\usepackage{float} % For better control over float placement

\definecolor{lightgray}{rgb}{0.95,0.95,0.95}

\usepackage{amsmath,amsfonts,bm}

\def\eqref#1{equation~\ref{#1}}
\def\1{\bm{1}}

\DeclareMathAlphabet{\mathsfit}{\encodingdefault}{\sfdefault}{m}{sl}
\SetMathAlphabet{\mathsfit}{bold}{\encodingdefault}{\sfdefault}{bx}{n}

\begin{document}

% Title and author information
\title{KohakuRAG: A simple RAG framework with hierarchical document indexing}

\author{
    \begin{tabular}{@{}c@{\hspace{2em}}c@{\hspace{2em}}c@{\hspace{2em}}c@{}}
    \textbf{Shih-Ying Yeh}$^{1,2,3,\dagger}$ & \textbf{Yueh-Feng Ku}$^{1}$ & \textbf{Ko-Wei Huang}$^{1}$ & \textbf{Buu-Khang Tu}$^{1}$
    \end{tabular}\\[0.8em]
    $^{1}$National Tsing Hua University \quad $^{2}$Comfy Org Research \quad $^{3}$Kohaku-Lab\\[0.3em]
    {\small $^{\dagger}$Corresponding author: \texttt{kohaku@kblueleaf.net} \quad $\bullet$ \quad Code: \url{https://github.com/KohakuBlueleaf/KohakuRAG}}
}

\maketitle

\begin{abstract}
Retrieval-augmented generation (RAG) systems that answer questions from document collections face compounding difficulties when high-precision citations are required: flat chunking strategies sacrifice document structure, single-query formulations miss relevant passages through vocabulary mismatch, and single-pass inference produces stochastic answers that vary in both content and citation selection. We present KohakuRAG, a hierarchical RAG framework that preserves document structure through a four-level tree representation (document $\rightarrow$ section $\rightarrow$ paragraph $\rightarrow$ sentence) with bottom-up embedding aggregation, improves retrieval coverage through an LLM-powered query planner with cross-query reranking, and stabilizes answers through ensemble inference with abstention-aware voting. We evaluate on the WattBot 2025 Challenge, a benchmark requiring systems to answer technical questions from 32 documents with $\pm$0.1\% numeric tolerance and exact source attribution. KohakuRAG achieves first place on both public and private leaderboards (final score 0.861), as the only team to maintain the top position across both evaluation partitions. Ablation studies reveal that prompt ordering (+80\% relative), retry mechanisms (+69\%), and ensemble voting with blank filtering (+1.2pp) each contribute substantially, while hierarchical dense retrieval alone matches hybrid sparse-dense approaches (BM25 adds only +3.1pp). We release KohakuRAG as open-source software at \url{https://github.com/KohakuBlueleaf/KohakuRAG}.
\end{abstract}

\section{Introduction}
\label{sec:introduction}

Large Language Models (LLMs) demonstrate remarkable capabilities across natural language processing tasks~\citep{brown2020language,openai2025gpt5}, yet suffer from hallucination, outdated knowledge, and inability to access domain-specific information absent from training data~\citep{huang2023hallucinationsurvey}. Retrieval-Augmented Generation (RAG) addresses these limitations by grounding LLM outputs in external knowledge sources~\citep{lewis2020retrieval,gao2024ragsurvey}.

\begin{figure}[t]
    \centering
    \includegraphics[width=\textwidth]{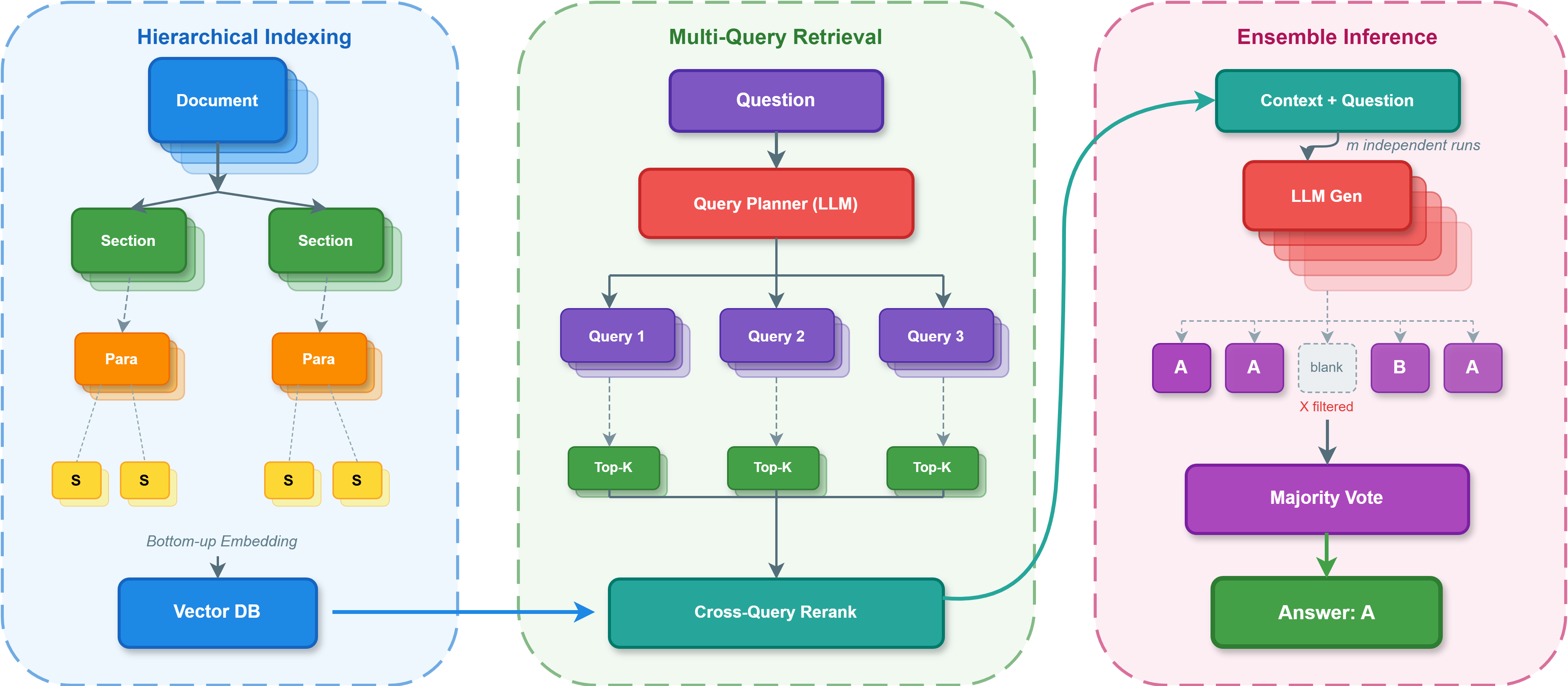}
    \caption{Overview of KohakuRAG. \textbf{Left (Hierarchical Indexing)}: Documents are parsed into tree structures with sections, paragraphs (Para), and sentences (S). Sentence embeddings are computed and aggregated bottom-up to parent levels, then stored in a Vector DB. \textbf{Center (Multi-Query Retrieval)}: Given a question, the Query Planner (LLM) generates multiple related queries, each retrieving Top-K results that are merged via Cross-Query Reranking. \textbf{Right (Ensemble Inference)}: Context and question are sent to the LLM for $m$ independent runs; blank responses are filtered (X), and majority voting produces the final answer.}
    \label{fig:overview}
\end{figure}

The WattBot 2025 Challenge~\citep{WattBot2025} presents a compelling testbed for RAG systems: given 32 reference documents (${\sim}$500K tokens) on AI energy consumption, systems must answer approximately 300 technical questions with $\pm$0.1\% numeric tolerance, cite exact source documents, and explicitly abstain when evidence is insufficient. This combination of scale, precision, and citation requirements exposes fundamental limitations of standard RAG approaches.

Standard RAG systems~\citep{lewis2020retrieval,huang2024ragsurvey} employ flat chunking that partitions documents into fixed-length segments, destroying structural boundaries and complicating precise citation tracking~\citep{chen2024latechunking}. Retrieval is typically performed with a single query formulation, which misses relevant passages when user questions use different vocabulary than source documents, e.g., querying ``PUE'' when documents discuss ``power usage effectiveness''~\citep{wang2023query2doc,jagerman2023queryexpansion}. Finally, a single LLM inference produces stochastic answers that vary across runs in both content and citation selection, with models often abstaining unnecessarily when evidence is present but difficult to locate.

We present KohakuRAG, a hierarchical RAG framework that addresses each of these challenges through a matched mechanism (Figure~\ref{fig:overview}):
\begin{itemize}[noitemsep,topsep=2pt]
    \item \textbf{Hierarchical document indexing.} To preserve document structure, we represent documents as four-level trees (document $\rightarrow$ section $\rightarrow$ paragraph $\rightarrow$ sentence) and compute embeddings bottom-up via length-weighted aggregation. This provides natural citation boundaries at each granularity level while capturing compositional semantics.
    \item \textbf{Multi-query retrieval with cross-query reranking.} To bridge the vocabulary gap between questions and source documents, an LLM-powered query planner expands questions into multiple retrieval queries covering alternative phrasings and terminology. Cross-query reranking then leverages consensus signals, where nodes retrieved by more queries rank higher, to surface the most relevant passages.
    \item \textbf{Ensemble inference with abstention-aware voting.} To mitigate answer instability, we aggregate $m$ independent inference runs via majority voting. By filtering out blank responses before aggregation, we prevent conservative runs from dominating when evidence exists but is hard to locate.
\end{itemize}

Our ablation studies quantify the impact of each design choice: prompt reordering (placing context before the question) yields +80\% relative improvement; the retry mechanism provides +69\% at low retrieval depth; and ensemble voting with blank filtering adds +1.2pp at $n$=9. Hierarchical dense retrieval alone achieves competitive performance, with BM25 augmentation adding only +3.1pp, suggesting that keyword matching provides diminishing returns when retrieval structure is sufficiently rich. KohakuRAG achieves first place on both the public and private leaderboards of the WattBot 2025 Challenge with a final score of 0.861. We were the only team to maintain the top position across both evaluation partitions, demonstrating that our ensemble-based approach generalizes reliably to unseen data.

Our main contributions are:
\begin{itemize}[noitemsep,topsep=2pt]
    \item A hierarchical document indexing scheme that preserves structural relationships through four-level tree representations with bottom-up embedding aggregation, enabling precise citation tracking.
    \item An LLM-powered query planner with cross-query reranking that improves retrieval coverage by expanding questions into multiple formulations and leveraging consensus signals.
    \item An ensemble inference mechanism with abstention-aware voting that aggregates multiple runs while filtering unnecessary abstentions, the dominant error mode (26.8\% of failures).
    \item Comprehensive experiments showing that hierarchical dense retrieval alone achieves competitive performance on citation-heavy tasks, and that prompt ordering and retry mechanisms contribute more than hybrid retrieval strategies.
\end{itemize}

The remainder of this paper is organized as follows. Section~\ref{sec:preliminaries} provides background on text retrieval and RAG systems. Section~\ref{sec:related_work} discusses related work. Section~\ref{sec:methodology} details our hierarchical indexing approach, retrieval strategies, and answering pipeline. Section~\ref{sec:experiment} describes our experimental setup, and Section~\ref{sec:results} presents results and analysis. Section~\ref{sec:conclusion} concludes with limitations and future directions.

\section{Preliminaries}
\label{sec:preliminaries}

We review the key research areas underlying our approach and formalize the WattBot 2025 Challenge.

\subsection{Large Language Models}

Large Language Models (LLMs) based on the transformer architecture~\citep{vaswani2017attention} have progressed from BERT~\citep{devlin2019bert} and GPT-3~\citep{brown2020language} to frontier models including GPT-5~\citep{openai2025gpt5}, Gemini 3~\citep{google2025gemini3}, Claude Opus 4.5~\citep{anthropic2025claude45}, and Grok 4.1~\citep{xai2025grok41}. Open-weight alternatives have also reached frontier quality: Mistral 3~\citep{mistral2025mistral3} with 675B total parameters and 41B active, Gemma 3~\citep{gemmateam2025gemma3}, and GPT-oss-120B~\citep{openai2025gptoss}. Despite strong reasoning capabilities, LLMs suffer from hallucination~\citep{huang2023hallucinationsurvey}, outdated knowledge, and inability to access domain-specific information, motivating retrieval-augmented approaches. Vision-language models (VLMs) extend LLMs with visual understanding: Qwen3-VL~\citep{bai2025qwen3vl} achieves state-of-the-art on MMMU and video understanding with native 256K-token multimodal context, while earlier work including LLaVA~\citep{liu2023llava} and Qwen-VL~\citep{bai2023qwenvl,wang2024qwen2vl} established key architectural patterns. Recent surveys~\citep{yin2024mllmsurvey} document rapid MLLM progress enabling document understanding where figures and tables contain critical information.

\subsection{Text Embeddings and Retrieval}

Dense retrieval represents queries and documents as vectors in $\mathbb{R}^d$, computing relevance as $\text{sim}(q, d) = E(q)^\top E(d)$ for encoder $E$. DPR~\citep{karpukhin2020dense} demonstrated that contrastive learning outperforms BM25~\citep{robertson2009probabilistic}, with improvements from Contriever~\citep{izacard2022contriever} and ANCE~\citep{xiong2021ance}. Modern embeddings achieve state-of-the-art on MTEB~\citep{muennighoff2023mteb} and its multilingual expansion MMTEB~\citep{mmteb2025}. Notable models include Sentence-BERT~\citep{reimers2019sentence}, GTE~\citep{li2023gte}, E5~\citep{wang2024e5}, BGE~\citep{xiao2024bge}, and the Jina series: Jina~v3~\citep{sturua2024jina} introduces task-specific LoRA adapters, while Jina~v4~\citep{gunther2025jina} unifies text and image embeddings through a 3.8B parameter multimodal architecture with Matryoshka representations~\citep{kusupati2022matryoshka} supporting dimensions from 128 to 2048.

\paragraph{Dense Vector Search.}
Efficient retrieval over large embedding collections requires approximate nearest neighbor (ANN) algorithms. HNSW~\citep{malkov2018hnsw} constructs hierarchical navigable small world graphs enabling logarithmic search complexity with high recall. The Faiss library~\citep{johnson2021faiss,douze2025faisslibrary} provides optimized implementations of multiple indexing strategies including inverted file indices (IVF), product quantization (PQ), and HNSW, supporting billion-scale search on both CPU and GPU. These algorithms trade exact retrieval for dramatic speedups, typically achieving $>$95\% recall while reducing latency by orders of magnitude compared to brute-force search.

\paragraph{Sparse Vector Search.}
Sparse methods represent documents as high-dimensional vectors with most entries zero, enabling efficient inverted index lookup. BM25~\citep{robertson2009probabilistic} remains a strong baseline, computing term-frequency statistics without training. Learned sparse methods like SPLADE~\citep{formal2021splade} use neural networks to learn term weights and expand documents with semantically related terms, combining the efficiency of sparse representations with learned relevance signals. Hybrid approaches that combine dense and sparse retrieval consistently outperform either method alone~\citep{chen2024sparsedense}, as lexical and semantic signals capture complementary relevance aspects.

Two-stage pipelines use fast first-stage retrieval followed by expensive reranking. Cross-encoders~\citep{nogueira2019passage} jointly encode query-document pairs with $O(n)$ cost; BGE-M3~\citep{xiao2024bgereranker} provides multi-granularity reranking. Late interaction models like ColBERT~\citep{khattab2020colbert,santhanam2022colbertv2} compute token-level MaxSim: $\text{score}(q, d) = \sum_i \max_j \mathbf{q}_i^\top \mathbf{d}_j$, balancing efficiency and accuracy. ColPali~\citep{faysse2024colpali} extends this to multimodal document retrieval over images without OCR.

\subsection{Retrieval-Augmented Generation}

RAG~\citep{lewis2020retrieval} conditions generation on retrieved documents: given corpus $\mathcal{D}$ and query $q$, retrieve $\mathcal{D}_q = \text{top-}k(\{d : \text{sim}(q, d)\})$, then generate $p(a | q, \mathcal{D}_q)$. Comprehensive surveys~\citep{wang2025ragsurvey,li2025agenticrag,gao2024ragsurvey,zhao2024ragaigc,huang2024ragsurvey} categorize systems as Na\"ive, Advanced (with query rewriting and reranking), Modular RAG, or Agentic RAG with autonomous retrieval decisions. Query expansion addresses vocabulary mismatch: Query2Doc~\citep{wang2023query2doc} generates pseudo-documents, while other approaches use rewriting~\citep{ma2023queryrewriting} or direct expansion~\citep{jagerman2023queryexpansion}. Document segmentation impacts quality; fixed chunking loses context, while late chunking~\citep{chen2024latechunking} encodes documents before segmentation. Our approach preserves natural structure as a tree $T = (V, E)$ with bottom-up embedding aggregation. For multimodal documents, M3DocRAG~\citep{cho2024m3docrag} shows combined text-visual retrieval improves performance. Self-consistency~\citep{wang2023selfconsistency} and ensemble methods~\citep{chen2024llmensemble} improve robustness via majority voting.

\subsection{Problem Formulation: The WattBot 2025 Challenge}

Document-grounded question answering spans several benchmarks with varying characteristics. Natural Questions~\citep{kwiatkowski2019naturalquestions} pairs real user queries with Wikipedia articles, requiring systems to extract short answers from long documents. TriviaQA~\citep{joshi2017triviaqa} provides trivia questions with multiple evidence documents, testing multi-hop reasoning. PDFTriage~\citep{talmor2019pdftriage} focuses on structured document QA with complex layouts. These benchmarks primarily evaluate answer extraction; citation accuracy and abstention behavior receive less attention.

The WattBot 2025 Challenge~\citep{WattBot2025} extends document-grounded QA to AI energy estimation, requiring systems to answer technical questions about power consumption, water usage, carbon footprint, and performance characteristics. We formalize the task as follows.

\paragraph{Task Definition.}
Given a document corpus $\mathcal{D} = \{d_1, \ldots, d_{32}\}$ and a question $q$, the system must produce a tuple $(a, R, b)$ where:
\begin{itemize}[noitemsep,topsep=2pt]
    \item $a \in \mathbb{R} \cup \mathcal{V}$ is the answer, either numeric or categorical from vocabulary $\mathcal{V}$
    \item $R \subseteq \{1, \ldots, 32\}$ is the set of cited document identifiers
    \item $b \in \{0, 1\}$ indicates abstention when evidence is insufficient
\end{itemize}
The corpus comprises 32 reference documents (research papers, technical reports, presentation slides) totaling approximately 500K tokens, exceeding typical LLM context limits and necessitating retrieval-based approaches.

\paragraph{Evaluation Protocol.}
The challenge imposes stringent requirements across three dimensions:

\textit{Answer Accuracy} (75\% weight): Numeric answers must fall within $\pm 0.1\%$ relative tolerance:
\begin{equation}
\text{ValueScore}(a, a^*) = \mathbf{1}\left[\left|\frac{a - a^*}{a^*}\right| \leq 0.001\right]
\label{eq:answer_tolerance}
\end{equation}
Categorical answers require exact string match after normalization.

\textit{Citation Precision} (15\% weight): Measured via Jaccard similarity between predicted and ground-truth reference sets:
\begin{equation}
\text{RefScore}(R, R^*) = \frac{|R \cap R^*|}{|R \cup R^*|}
\label{eq:ref_score}
\end{equation}

\textit{Hallucination Avoidance} (10\% weight): Systems must explicitly abstain ($b=1$) rather than hallucinate when evidence is insufficient. Incorrect answers when abstention was appropriate incur penalties.

The final score combines these components:
\begin{equation}
\text{Score} = 0.75 \times \text{ValueScore} + 0.15 \times \text{RefScore} + 0.10 \times \text{HallucinationScore}
\label{eq:final_score}
\end{equation}
requiring systems to jointly optimize retrieval coverage, answer extraction, and uncertainty quantification.

\section{Related Work}
\label{sec:related_work}

We discuss prior work on document chunking strategies, query expansion methods, and ensemble approaches for RAG systems.

\paragraph{Document Chunking and Hierarchical Indexing.}
Standard RAG systems partition documents into fixed-length chunks for embedding and retrieval~\citep{lewis2020retrieval,gao2024ragsurvey}. While simple to implement, fixed-length chunking destroys semantic boundaries and complicates citation tracking. Late chunking~\citep{chen2024latechunking} addresses boundary artifacts by encoding documents before segmentation, preserving contextual information across chunk boundaries. However, late chunking still produces flat representations without explicit hierarchical structure. Our approach differs by constructing explicit tree representations that preserve document organization (sections, paragraphs, sentences) and enable precise provenance tracking through hierarchical node identifiers.

\paragraph{Query Expansion and Multi-Query Retrieval.}
Single queries often lack sufficient context for optimal retrieval, motivating query expansion techniques. Query2Doc~\citep{wang2023query2doc} generates pseudo-documents to enrich query representations. Other approaches employ query rewriting~\citep{ma2023queryrewriting} or direct expansion through prompting~\citep{jagerman2023queryexpansion}. We adopt LLM-powered query planning that generates multiple semantically related queries, combined with cross-query reranking that leverages consensus signals, where nodes retrieved by multiple queries rank higher, providing implicit voting across different query formulations.

\paragraph{Self-Reflection and Corrective RAG.}
Recent work explores mechanisms for RAG systems to assess and improve their own outputs. Self-RAG~\citep{asai2023selfrag} trains models to generate special reflection tokens that indicate when to retrieve, whether retrieved passages are relevant, and whether generated content is supported by evidence. CRAG~\citep{yan2024crag} introduces a corrective retrieval mechanism that evaluates retrieval quality and triggers web search when initial retrieval is insufficient. Our retry mechanism shares similar motivation: when the model indicates insufficient evidence (abstention), we expand context and retry. However, our approach operates at inference time without requiring specialized training.

\paragraph{Ensemble Methods for RAG.}
Self-consistency~\citep{wang2023selfconsistency} improves reasoning by sampling multiple outputs and selecting the most common answer through majority voting. Recent surveys on LLM ensembles~\citep{chen2024llmensemble} document various aggregation strategies for combining predictions across models or runs. Our ensemble mechanism extends these ideas to RAG with reference-aware voting modes that jointly consider answer values and citation sets, with explicit handling of abstention cases where some runs indicate insufficient evidence.

\paragraph{Multimodal Document Understanding.}
Technical documents frequently contain critical information in figures and tables. M3DocRAG~\citep{cho2024m3docrag} demonstrates that combining text and visual retrieval improves performance on multi-page document understanding. ColPali~\citep{faysse2024colpali} enables document retrieval directly from page images using late interaction over vision embeddings. Our framework supports both caption-based retrieval (where vision models generate textual descriptions during indexing) and direct image embedding through Jina v4~\citep{gunther2025jina}, enabling flexible multimodal retrieval strategies.

\section{Methodology}
\label{sec:methodology}

We present KohakuRAG, a hierarchical RAG framework designed for document-grounded question answering with precise citation requirements. Our approach addresses three key challenges: (1) preserving document structure during indexing, (2) improving retrieval coverage through multi-query planning, and (3) achieving robust answers via ensemble inference. Algorithm~\ref{alg:kohakurag} presents the complete pipeline.

\begin{figure}[t]
\begin{minipage}[t]{0.48\textwidth}
\vspace{0pt}
The KohakuRAG pipeline operates in three phases. First, given question $q$, an LLM-powered query planner generates $n$ semantically related queries $\mathcal{Q} = \{q_1, \ldots, q_n\}$ that cover different phrasings and aspects of the information need. Each query retrieves top-$k$ nodes via dense similarity search, and results are aggregated through cross-query reranking that prioritizes nodes retrieved by multiple queries.

Second, retrieved nodes are expanded with hierarchical context, including parent paragraphs and sibling sentences, to provide the LLM with sufficient surrounding information.

Third, $m$ independent inference runs generate answer candidates $(a^{(j)}, R^{(j)}, b^{(j)})$ where $b^{(j)}$ indicates abstention. If $b^{(j)}=1$ and retries remain, we increase $k$ and re-retrieve. Finally, ensemble voting with blank filtering produces the final answer.
\end{minipage}%
\hfill
\begin{minipage}[t]{0.50\textwidth}
\vspace{0pt}
\begin{algorithm}[H]
\caption{KohakuRAG Pipeline}
\label{alg:kohakurag}
\begin{algorithmic}[1]
\small
\Require $q$, index $\mathcal{I}$, $k$, $n_{\text{queries}}$, $m_{\text{runs}}$
\Ensure $(a, R, b)$
\Function{KohakuRAG}{$q, \mathcal{I}, k, n, m$}
    \State $\mathcal{Q} \gets \textsc{PlanQueries}(q, n)$
    \State $\mathcal{R} \gets \emptyset$
    \For{$q_i \in \mathcal{Q}$}
        \State $\mathcal{R} \gets \mathcal{R} \cup \textsc{Retrieve}(\mathcal{I}, q_i, k)$
    \EndFor
    \State $\mathcal{R} \gets \textsc{Rerank}(\mathcal{R})$
    \State $\mathcal{C} \gets \textsc{ExpandContext}(\mathcal{R})$
    \State $\mathcal{A} \gets \emptyset$
    \For{$j = 1$ \textbf{to} $m$}
        \State $(a^{(j)}, R^{(j)}, b^{(j)}) \gets \textsc{Generate}(\mathcal{C}, q)$
        \If{$b^{(j)} = 1$ \textbf{and} retries left}
            \State $k \gets k + \Delta k$; \textbf{goto} line 4
        \EndIf
        \State $\mathcal{A} \gets \mathcal{A} \cup \{(a^{(j)}, R^{(j)}, b^{(j)})\}$
    \EndFor
    \State \Return $\textsc{EnsembleVote}(\mathcal{A})$
\EndFunction
\end{algorithmic}
\end{algorithm}
\end{minipage}
\end{figure}

\subsection{Hierarchical Document Indexing}
\label{subsec:hierarchical_indexing}

Standard RAG systems employ flat chunking strategies that partition documents into fixed-length segments. While simple, this approach destroys document structure and complicates citation tracking. We instead preserve the natural hierarchy of technical documents through tree-structured indexing.

\paragraph{Document Parsing.}
Given a document $d$, we construct a tree $T_d = (V_d, E_d)$ with four levels of granularity. The root node represents the entire document. Section nodes correspond to logical divisions (chapters, numbered sections). Paragraph nodes capture coherent text blocks within sections. Sentence nodes form the leaves, containing individual sentences extracted via rule-based segmentation. Each node $v$ stores its text content $t_v$, a unique identifier following the pattern \texttt{doc:sec$i$:p$j$:s$k$}, and metadata including source document ID and position indices.

For documents containing figures and tables, we treat each visual element as a special paragraph node with \texttt{attachment\_type=image}. We employ a Vision Language Model (VLM) from the Qwen-VL family~\citep{bai2023qwenvl,bai2025qwen3vl} to generate captions, which are stored as the node's text content. This enables text-based retrieval of visual information while preserving the hierarchical structure.

\paragraph{Bottom-Up Embedding Propagation.}
We compute embeddings in a bottom-up manner to capture compositional semantics. Let $\mathcal{C}(v)$ denote the children of node $v$. For leaf nodes (sentences), we directly compute embeddings using a text encoder $E$:
\begin{equation}
\mathbf{e}_v = E(t_v), \quad \text{if } \mathcal{C}(v) = \emptyset
\end{equation}

For internal nodes, the embedding is computed as a weighted average of child embeddings:
\begin{equation}
\mathbf{e}_v = \frac{\sum_{c \in \mathcal{C}(v)} w_c \cdot \mathbf{e}_c}{\sum_{c \in \mathcal{C}(v)} w_c}
\label{eq:embedding_propagation}
\end{equation}
where $w_c = |t_c|$ is the token count of child $c$. This length-weighted averaging ensures that longer, more informative children contribute proportionally more to the parent embedding. The resulting embeddings at paragraph and section levels capture the aggregate semantics of their constituents.

\paragraph{Embedding Models.}
We support two embedding backends. Jina Embeddings v3~\citep{sturua2024jina} provides 768-dimensional text embeddings with 8K context length and task-specific LoRA adapters, trained with multi-stage contrastive learning similar to GTE~\citep{li2023gte}, E5~\citep{wang2024e5}, and BGE~\citep{xiao2024bge}. These models achieve state-of-the-art results on the MTEB benchmark~\citep{muennighoff2023mteb}. Jina Embeddings v4~\citep{gunther2025jina} extends this with multimodal capability and Matryoshka representations~\citep{kusupati2022matryoshka} (128--2048 dimensions), enabling direct image embedding for cross-modal retrieval.

\paragraph{Vector Storage.}
Embeddings are stored using sqlite-vec, a lightweight SQLite extension for vector similarity search. Our middleware layer, KohakuVault, provides a unified interface for key-value metadata storage and vector indexing within a single database file. This design enables efficient deployment without external database dependencies.

\subsection{Multi-Query Retrieval}
\label{subsec:retrieval}

User questions often lack sufficient context for optimal retrieval. A query like ``What is the PUE of Google's data centers?'' may miss relevant passages discussing ``power usage effectiveness'' or ``energy efficiency metrics.'' We address this through LLM-powered query planning.

\paragraph{Query Planner.}
Given question $q$, an LLM generates $n$ semantically related queries $\{q_1, \ldots, q_n\}$ that cover different phrasings and aspects of the information need. The planner is prompted to produce queries that:
\begin{itemize}[noitemsep,topsep=0pt]
    \item Rephrase the original question using alternative terminology
    \item Expand abbreviations and technical terms
    \item Decompose compound questions into sub-queries
    \item Add contextual keywords likely to appear in relevant passages
\end{itemize}

\paragraph{Dense Retrieval.}
For each planned query $q_i$, we embed it and retrieve the top-$k$ matching nodes via cosine similarity search:
\begin{equation}
\mathcal{R}_i = \text{top-}k\left(\{v \in V : \cos(\mathbf{e}_{q_i}, \mathbf{e}_v)\}, k\right)
\end{equation}
Retrieval is restricted to sentence and paragraph nodes, as these contain the most specific information. Section and document nodes, while indexed, are too coarse for precise matching.

\paragraph{Cross-Query Reranking.}
Results from multiple queries are aggregated and reranked. While neural rerankers such as cross-encoders~\citep{nogueira2019passage,xiao2024bgereranker} and late interaction models~\citep{khattab2020colbert,santhanam2022colbertv2} offer strong performance, we employ a lightweight score-based reranking to minimize latency. For each unique node $v$ appearing in $\bigcup_i \mathcal{R}_i$, we compute:
\begin{itemize}[noitemsep,topsep=0pt]
    \item \textbf{Frequency}: $f_v = |\{i : v \in \mathcal{R}_i\}|$, the number of queries that retrieved $v$
    \item \textbf{Total score}: $s_v = \sum_{i : v \in \mathcal{R}_i} \text{sim}(q_i, v)$, the cumulative similarity
\end{itemize}

We support three reranking strategies:
\begin{itemize}[noitemsep,topsep=0pt]
    \item \textsc{Frequency}: Sort by $(f_v, s_v)$ descending, so nodes retrieved by more queries rank higher
    \item \textsc{Score}: Sort by $s_v$ descending, so nodes with highest cumulative similarity rank higher
    \item \textsc{Combined}: Sort by $\alpha \cdot \hat{f}_v + (1-\alpha) \cdot \hat{s}_v$ where $\hat{\cdot}$ denotes min-max normalization
\end{itemize}

After reranking, results are optionally truncated to $k_{\text{final}}$ nodes to control context length.

\paragraph{Hierarchical Context Expansion.}
Retrieved nodes are expanded with their hierarchical context. For each matched node $v$, we include its parent node (providing broader context) and sibling sentences (providing local context). This expansion leverages the tree structure: a matched sentence automatically brings its containing paragraph, enabling the LLM to understand the surrounding discussion without redundant retrieval. Since multi-query retrieval often returns the same node from different queries, we deduplicate by node ID after reranking, removing cross-query duplicates before context expansion.

\begin{figure}[t]
    \centering
    \includegraphics[width=\textwidth]{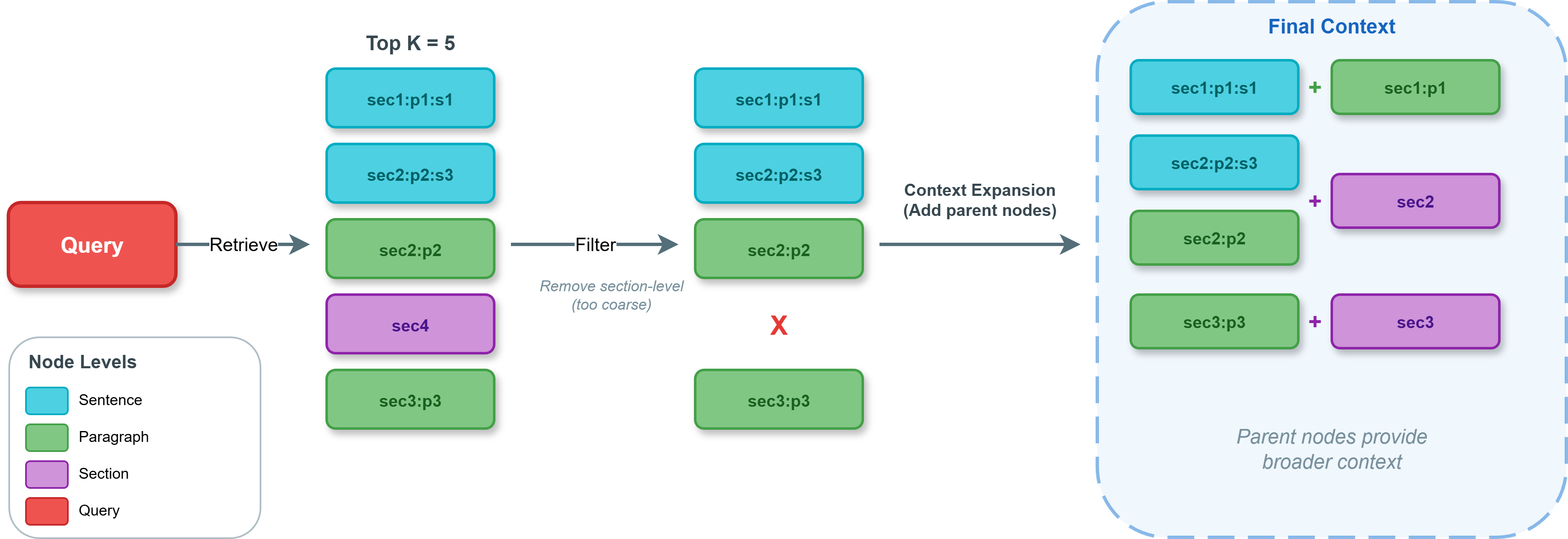}
    \caption{Hierarchical context expansion. Given a query, we retrieve Top-K nodes across multiple granularity levels (sentences in cyan, paragraphs in green, sections in purple). Section-level nodes are filtered out as too coarse (marked with X). The remaining nodes are expanded by adding their parent nodes to provide broader context. For example, a matched sentence \texttt{sec1:p1:s1} brings its parent paragraph \texttt{sec1:p1}, while a matched paragraph \texttt{sec2:p2} brings its parent section \texttt{sec2}.}
    \label{fig:context_expansion}
\end{figure}

\paragraph{Optional BM25 Augmentation.}
For queries requiring exact keyword matching (e.g., specific model names or numerical values), we optionally augment dense retrieval with BM25~\citep{robertson2009probabilistic} results. While learned sparse methods like SPLADE~\citep{formal2021splade} offer improved performance, we use traditional BM25 for simplicity. Hybrid sparse-dense retrieval has been shown to provide complementary benefits~\citep{chen2024sparsedense}; the sparse results are appended after dense results without score fusion, providing complementary lexical matches.

\subsection{Answering Pipeline}
\label{subsec:answering}

\begin{figure}[t]
    \centering
    \includegraphics[width=\textwidth]{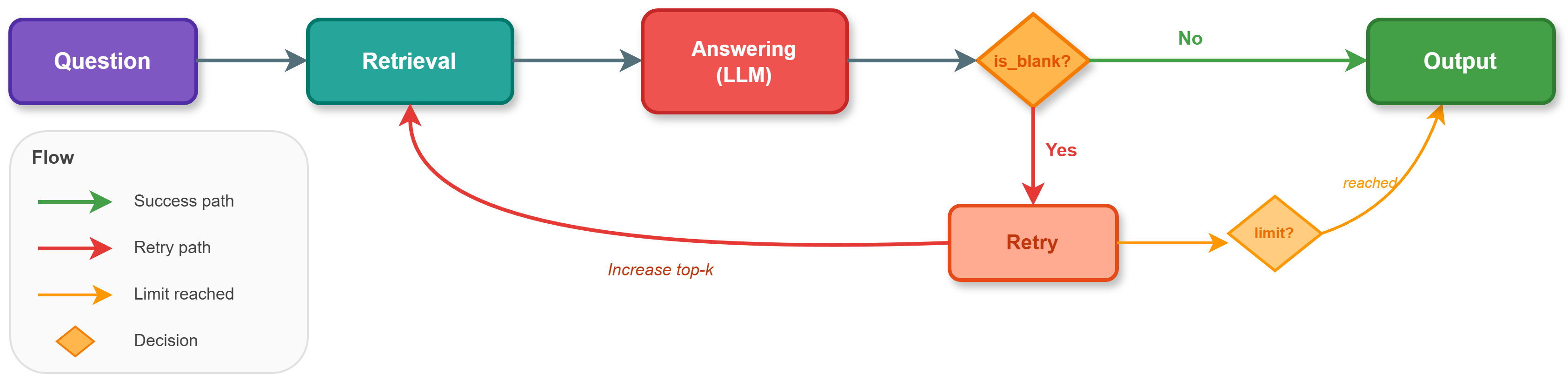}
    \caption{Retry mechanism flow. When the LLM outputs \texttt{is\_blank=true} (insufficient evidence), the system increases top-$k$ and re-retrieves context. This loop continues until a valid answer is produced or the retry limit is reached, at which point the system outputs a blank response.}
    \label{fig:retry_mechanism}
\end{figure}

\paragraph{Prompt Construction.}
Retrieved snippets are formatted with explicit reference markers: \texttt{[ref\_id=\{doc\_id\}] \{text\}}. This format enables the LLM to directly cite sources by document ID. For documents with images, captions are included in a separate ``Referenced media'' section.

We found that prompt ordering significantly affects answer quality. Placing the question and instructions \textit{after} the context (rather than before) yields better results, consistent with the ``lost in the middle'' phenomenon~\citep{liu2024lost}. This effect arises from attention sink behavior~\citep{xiao2024efficientstreaminglanguagemodels}, where initial tokens accumulate disproportionate attention mass due to softmax normalization, combined with recency bias from causal attention and rotary position encodings~\citep{su2023roformerenhancedtransformerrotary}.

\paragraph{Structured Output.}
The LLM is instructed to output JSON with four fields:
\begin{itemize}[noitemsep,topsep=0pt]
    \item \texttt{answer}: Natural language response
    \item \texttt{answer\_value}: Extracted value (numeric or categorical)
    \item \texttt{ref\_id}: List of source document IDs
    \item \texttt{explanation}: Reasoning chain (for debugging)
\end{itemize}
When evidence is insufficient, the model outputs \texttt{is\_blank=true} to explicitly abstain rather than hallucinate.

\paragraph{Retry Mechanism.}
If the LLM indicates insufficient evidence (Figure~\ref{fig:retry_mechanism}), we retry with increased $k$ to provide more context. This iterative deepening allows the system to find relevant information that may have ranked just below the initial cutoff.

\subsection{Ensemble Inference}
\label{subsec:ensemble}

Single-inference answers exhibit variance due to LLM sampling stochasticity. Following self-consistency~\citep{wang2023selfconsistency} and recent work on LLM ensembles~\citep{chen2024llmensemble}, we improve robustness through ensemble inference with majority voting.

\paragraph{Multi-Run Aggregation.}
For each question, we perform $m$ independent inference runs with temperature $>0$. Each run produces an answer tuple $(a^{(j)}, R^{(j)})$ for $j \in \{1, \ldots, m\}$.

\paragraph{Voting Strategies.}
We support multiple voting modes:
\begin{itemize}[noitemsep,topsep=0pt]
    \item \textsc{Independent}: Vote on answer and references separately
    \item \textsc{AnswerPriority}: First vote on answer, then collect references from matching runs
    \item \textsc{RefPriority}: First vote on references, then select answer from matching runs
    \item \textsc{Union}: Vote on answer, take union of references from matching runs
    \item \textsc{Intersection}: Vote on answer, take intersection of references
\end{itemize}

\begin{figure}[t]
    \centering
    \includegraphics[width=\textwidth]{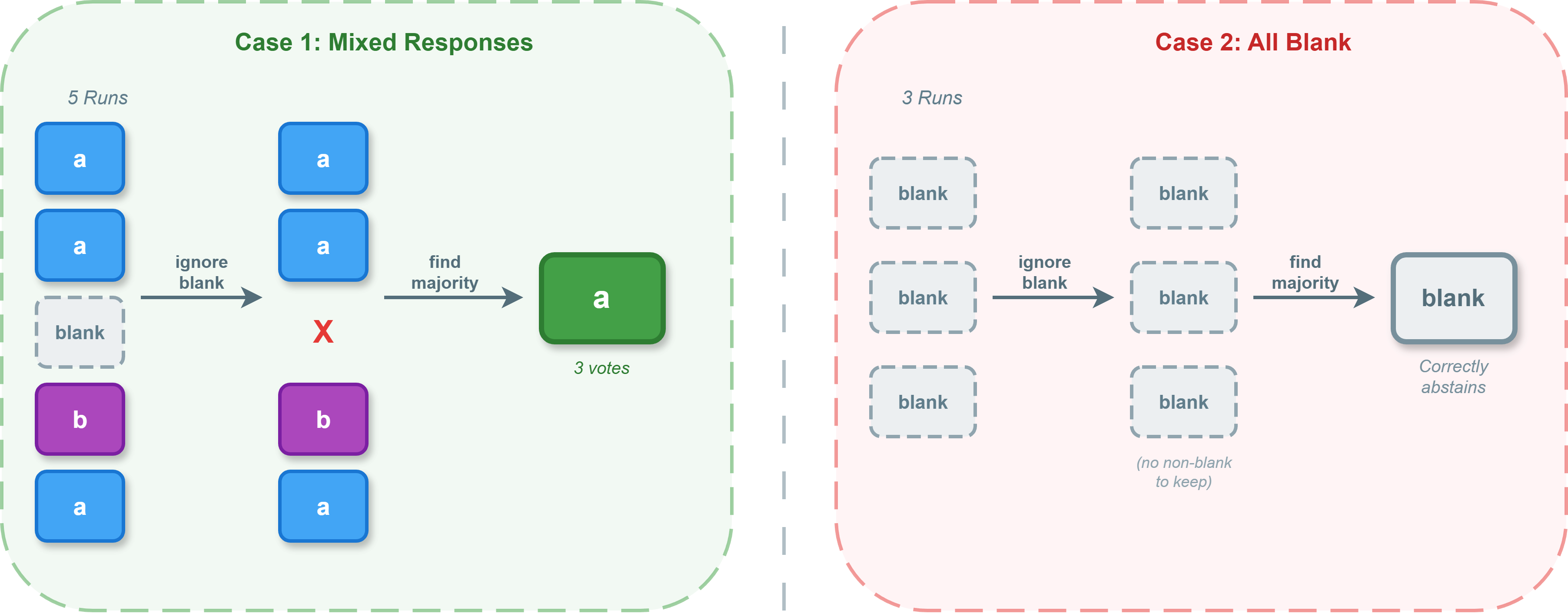}
    \caption{Ensemble voting with blank handling. \textbf{Case 1 (Mixed Responses)}: When non-blank answers exist among 5 runs \{a, a, blank, b, a\}, blank responses are filtered (X) before majority voting, yielding \{a, a, b, a\} with majority ``a'' (3 votes). \textbf{Case 2 (All Blank)}: When all runs return blank, the system correctly abstains rather than hallucinating, preserving the hallucination score.}
    \label{fig:ensemble_voting}
\end{figure}

\paragraph{Blank Handling.}
A critical design choice is handling abstention (\texttt{is\_blank}) answers. Setting \texttt{ignore\_blank=true} filters out blank answers before voting if any non-blank answer exists. This prevents conservative runs from dominating when evidence exists but is difficult to find. Figure~\ref{fig:ensemble_voting} illustrates the two cases.

\subsection{Multimodal Retrieval}
\label{subsec:multimodal}

\begin{figure}[t]
\begin{minipage}[t]{0.52\textwidth}
\vspace{0pt}
Technical documents often contain critical information in figures and tables~\citep{cho2024m3docrag,yin2024mllmsurvey}. KohakuRAG supports two modes of multimodal retrieval:

\textbf{Caption-Based Retrieval.}
Image captions generated during indexing are stored as paragraph nodes. Text queries naturally retrieve relevant images through their captions, which are included in the LLM prompt as text.

\textbf{Vision-Enabled Retrieval.}
With Jina v4 embeddings, images can be directly embedded and retrieved, similar to ColPali~\citep{faysse2024colpali}. For vision-capable LLMs such as GPT-4V, Gemini, and Qwen2-VL~\citep{wang2024qwen2vl,liu2023llava}, retrieved images are sent as base64-encoded content alongside text, enabling direct interpretation of visual information.
\end{minipage}%
\hfill
\begin{minipage}[t]{0.46\textwidth}
\vspace{0pt}
\centering
\includegraphics[width=\textwidth]{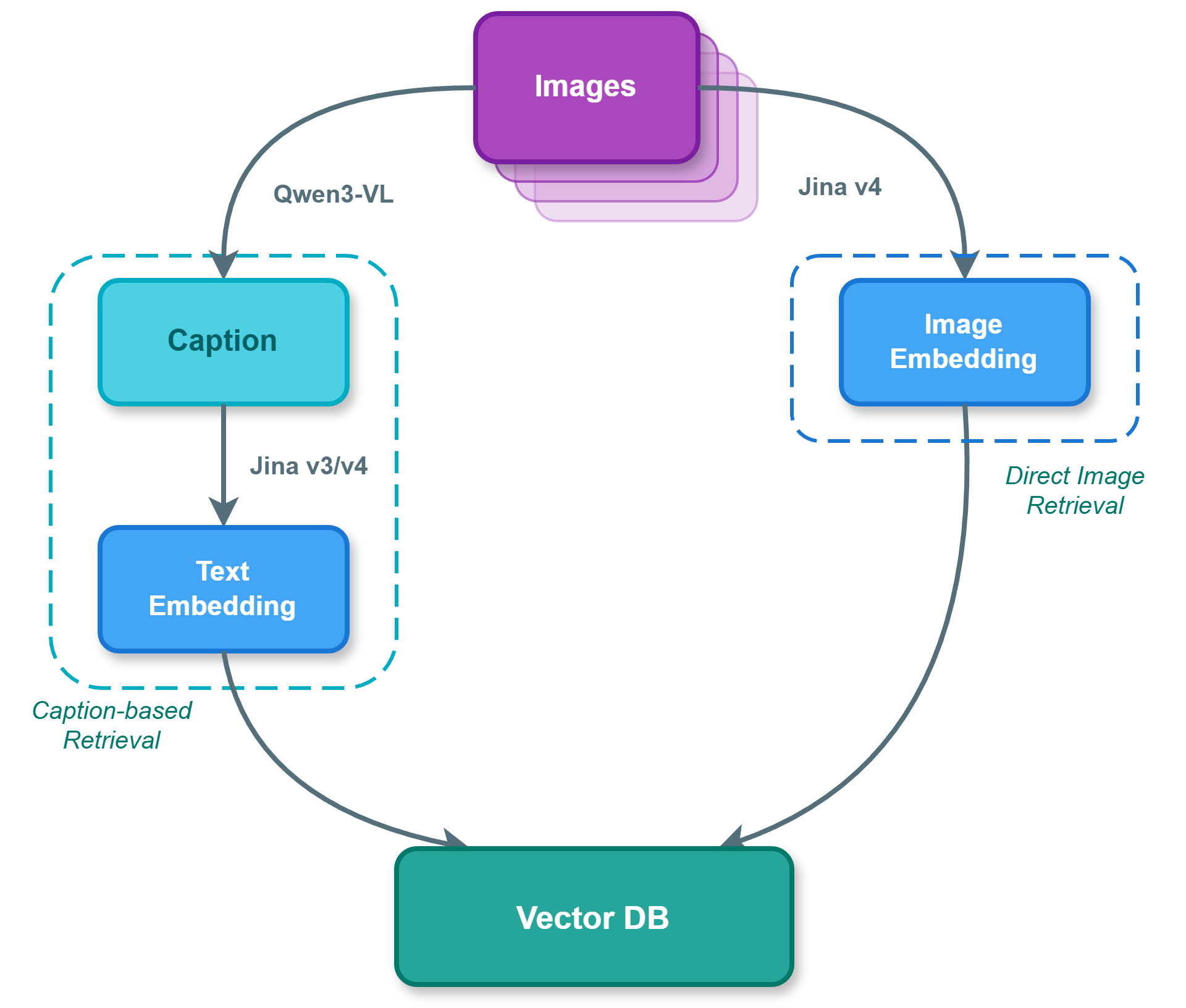}
\captionof{figure}{Dual-path image processing. \textbf{Left}: Caption-based retrieval via Qwen3-VL captioning and Jina v3/v4 text embedding. \textbf{Right}: Direct image embedding via Jina v4's multimodal encoder.}
\label{fig:image_processing}
\end{minipage}
\end{figure}

\section{Experiments}
\label{sec:experiment}

We conduct ablation studies to evaluate each component of KohakuRAG on the WattBot 2025 Challenge training set.

\subsection{Experimental Settings}
\label{subsec:setup}

\paragraph{Dataset and Corpus.}
The WattBot 2025 corpus comprises 32 reference documents (research papers and technical slides) totaling approximately 500K tokens. The training set contains 41 questions with ground-truth answers and reference IDs, spanning AI energy characteristics including power consumption, water usage, carbon footprint, and performance metrics. After hierarchical parsing, our index contains 32 documents, 639 sections, 944 paragraphs, 24,565 sentences, and 280 images (275 with extractable content).

\begin{figure}[t]
    \centering
    \includegraphics[width=\textwidth]{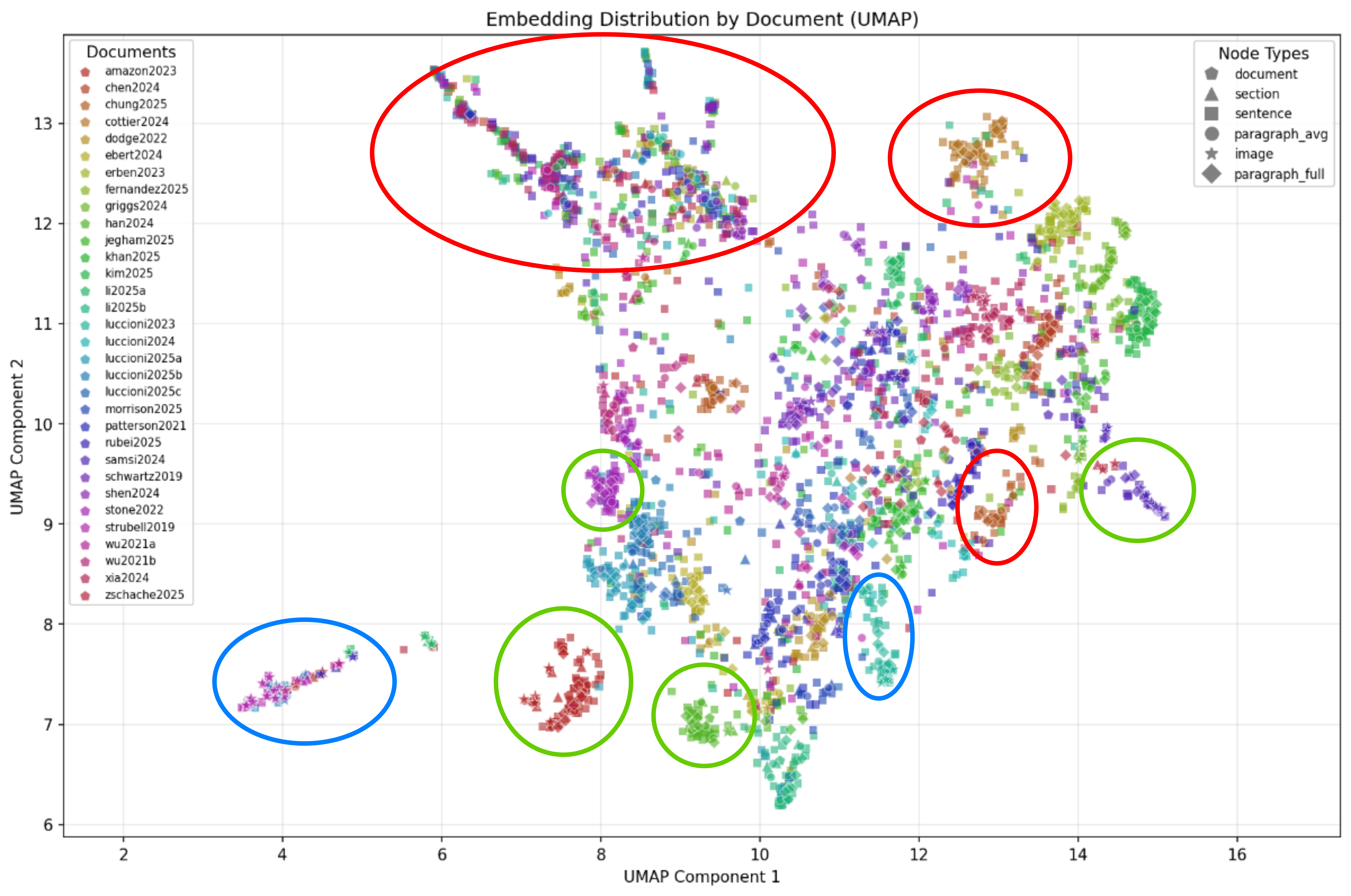}
    \caption{UMAP visualization of Jina v4 embeddings for all indexed nodes, colored by source document. Points are differentiated by node type (documents, sections, sentences, paragraph averages, images, and full paragraphs). We annotate three types of regions: \textcolor{green}{\textbf{green circles}} mark document-dominant clusters where one paper dominates 50--99\% of nodes with minor contributions from related documents; \textcolor{red}{\textbf{red circles}} highlight cross-document semantic clusters where multiple documents mix, indicating shared topics (e.g., energy measurement methodologies); \textcolor{blue}{\textbf{blue circles}} show image-text alignment regions where image embeddings (stars) cluster with semantically related sentences. This validates that our hierarchical embeddings preserve document coherence while enabling cross-document similarity matching.}
    \label{fig:umap_embedding}
\end{figure}

\paragraph{Evaluation Metrics.}
Following the official WattBot scoring protocol: \textbf{Value Score} (75\%) measures answer accuracy with $\pm 0.1\%$ tolerance for numerics; \textbf{Reference Score} (15\%) computes Jaccard similarity of reference IDs; \textbf{Hallucination Score} (10\%) rewards correct abstention. Final score: $0.75 \times \text{value} + 0.15 \times \text{ref} + 0.10 \times \text{hallucination}$.

\paragraph{Default Configuration.}
Unless otherwise noted, we use GPT-oss-120B as the generator, Jina v3 embeddings (768-dim), top-$k$=16 per query, 4 planner queries, combined reranking, and top-$k_{\text{final}}$=32. Each configuration is evaluated over 3 runs with different random seeds. Statistical significance is assessed using Welch's $t$-test (unequal variances); we report $p$$<$0.05 as significant and $p$$<$0.10 as marginally significant given the inherent stochasticity of LLM sampling.

\subsection{Retrieval Component Analysis}
\label{subsec:retrieval_analysis}

We evaluate embedding models, query planning, and reranking strategies. Table~\ref{tab:retrieval} presents results across retrieval depths $k \in \{4, 8, 16\}$.

\begin{table}[t]
\centering
\caption{Retrieval component ablations (final score). Best per column in \textbf{bold}, second best \underline{underlined}. Significance vs.\ best: $^{\dagger}$$p$$<$0.10, $^{*}$$p$$<$0.05, $^{**}$$p$$<$0.01.}
\label{tab:retrieval}
\small
\begin{tabular}{llcccc}
\toprule
\textbf{Component} & \textbf{Setting} & \textbf{$k$=4} & \textbf{$k$=8} & \textbf{$k$=16} & \textbf{Avg Rank} \\
\midrule
\multirow{3}{*}{Embedding}
    & Jina v3 (text only) & 0.752$^{**}$ & 0.801$^{\dagger}$ & 0.824$^{*}$ & 3.00 \\
    & Jina v3 (+ captions) & \underline{0.859} & \underline{0.836}$^{\dagger}$ & \underline{0.824}$^{**}$ & 2.00 \\
    & Jina v4 (+ images) & \textbf{0.869} & \textbf{0.868} & \textbf{0.890} & \textbf{1.00} \\
\midrule
\multirow{3}{*}{Planner}
    & 2 queries & 0.562$^{**}$ & 0.466$^{*}$ & 0.558$^{**}$ & 3.00 \\
    & 4 queries & \underline{0.678} & \textbf{0.830} & \underline{0.811} & 1.67 \\
    & 6 queries & \textbf{0.787} & \underline{0.809} & \textbf{0.838} & \textbf{1.33} \\
\midrule
\multirow{4}{*}{Reranking}
    & None & \textbf{0.808} & 0.765 & \textbf{0.833} & \textbf{1.67} \\
    & Frequency & 0.396$^{*}$ & 0.533$^{**}$ & 0.585$^{**}$ & 4.00 \\
    & Score & 0.605 & \underline{0.792} & 0.785$^{*}$ & 2.67 \\
    & Combined & \underline{0.798} & \textbf{0.822} & \underline{0.814} & \textbf{1.67} \\
\bottomrule
\end{tabular}
\end{table}

Jina v4 with native image embeddings achieves the best retrieval quality (0.890 at $k$=16), outperforming text-only Jina v3 by 6.6 percentage points ($p$=0.016, paired $t$-test). Increasing planner queries from 2 to 6 improves performance substantially (0.562$\rightarrow$0.787 at $k$=4, +40\% relative), as diverse query formulations capture different aspects of the information need. For reranking, the \textsc{Combined} strategy provides the most consistent results across retrieval depths, while \textsc{Frequency}-only yields substantially lower scores (0.396 vs 0.808 at $k$=4; $p$$<$0.01), demonstrating that retrieval scores contain important complementary information beyond frequency signals.

\subsection{Context Enhancement}
\label{subsec:context}

Table~\ref{tab:context} evaluates prompt reordering, BM25 hybrid retrieval, and the retry mechanism.

\begin{table}[t]
\centering
\caption{Context enhancement ablations (final score). Best per column in \textbf{bold}, second best \underline{underlined}. Significance vs.\ best: $^{\dagger}$$p$$<$0.10, $^{*}$$p$$<$0.05, $^{**}$$p$$<$0.01, $^{***}$$p$$<$0.001.}
\label{tab:context}
\small
\begin{tabular}{llcccc}
\toprule
\textbf{Component} & \textbf{Setting} & \textbf{$k$=4} & \textbf{$k$=8} & \textbf{$k$=16} & \textbf{Avg Rank} \\
\midrule
\multirow{2}{*}{Prompt Order}
    & Standard (Q$\rightarrow$C) & 0.418$^{*}$ & 0.485$^{***}$ & \underline{0.704} & 2.00 \\
    & Reordered (C$\rightarrow$Q) & \textbf{0.752} & \textbf{0.783} & \textbf{0.811} & \textbf{1.00} \\
\midrule
\multirow{4}{*}{BM25 top-$k$}
    & 0 (dense only) & 0.688$^{*}$ & 0.741 & 0.788 & 3.67 \\
    & 2 & 0.701 & \textbf{0.822} & \underline{0.812} & 2.00 \\
    & 4 & \textbf{0.763} & \underline{0.801} & \textbf{0.819} & \textbf{1.33} \\
    & 8 & \underline{0.720} & 0.720$^{**}$ & 0.806 & 3.00 \\
\midrule
\multirow{3}{*}{Max Retries}
    & 0 (disabled) & 0.488$^{**}$ & \underline{0.836} & \underline{0.837} & 3.00 \\
    & 1 & \underline{0.720} & 0.843 & 0.846 & 2.00 \\
    & 2 & \textbf{0.827} & \textbf{0.854} & \textbf{0.871} & \textbf{1.00} \\
\bottomrule
\end{tabular}
\end{table}

Prompt reordering yields substantial improvements (0.418$\rightarrow$0.752 at $k$=4, +80\% relative), confirming the ``lost in the middle'' phenomenon~\citep{liu2024lost}. Adding BM25 improves dense-only retrieval by 3.1 percentage points at $k$=16 (0.788$\rightarrow$0.819). The retry mechanism provides +69\% relative improvement at $k$=4 (0.488$\rightarrow$0.827), effectively addressing abstention failures where the model incorrectly abstains due to insufficient initial context.

\subsection{LLM Comparison}
\label{subsec:llm}

We compare seven LLMs using Jina v4 embeddings, including open-weight models like GPT-oss-120B~\citep{openai2025gptoss}. Table~\ref{tab:llm} reports mean scores across 3 runs.

\begin{table}[h]
\centering
\caption{LLM comparison (final score). Best in \textbf{bold}, second \underline{underlined}. Significance: $^{\dagger}$$p$$<$0.10, $^{*}$$p$$<$0.05, $^{**}$$p$$<$0.01.}
\label{tab:llm}
\small
\begin{tabular}{lcccc}
\toprule
\textbf{Model} & $k$=4 & $k$=8 & $k$=16 & \textbf{Rank} \\
\midrule
GPT-5-nano & .783$^{**}$ & .746$^{**}$ & .787$^{**}$ & 7.00 \\
GPT-5-mini & .812$^{*}$ & .804$^{*}$ & .847 & 4.67 \\
GPT-oss-120B & .805$^{\dagger}$ & .763$^{*}$ & .833$^{*}$ & 5.67 \\
Mistral-large & .853$^{*}$ & .862 & .816$^{\dagger}$ & 4.67 \\
Gemini-3-pro & .871$^{\dagger}$ & .873 & \underline{.875} & 2.67 \\
Kimi-k2 & \underline{.872}$^{*}$ & \textbf{.889} & .874 & 2.00 \\
Grok-4.1-fast & \textbf{.908} & \underline{.885} & \textbf{.888} & \textbf{1.33} \\
\bottomrule
\end{tabular}
\end{table}

Grok-4.1-fast achieves the best average rank (1.33) across all retrieval depths, with peak performance of 0.908 at $k$=4. The gap between Grok-4.1-fast and GPT-5-nano is highly significant ($p$$<$0.001), while differences among top-tier models are not statistically significant ($p$$>$0.1). Gemini-3-pro uniquely supports native image input. Smaller models (GPT-5-nano, avg rank 7.00) show significant degradation, indicating model capacity is critical.

\subsection{Ensemble Inference}
\label{subsec:ensemble_analysis}

Table~\ref{tab:ensemble} presents ensemble experiments with varying sizes and voting strategies.

\begin{table}[t]
\caption{Ensemble analysis. Left: ignore\_blank ablation (avg rank in parentheses); all differences for $n$$\geq$3 are significant ($p$$<$0.01). Right: voting strategy comparison at $n$=15; differences among top strategies are not significant ($p$$>$0.1).}
\label{tab:ensemble}
\centering
\small
\begin{tabular}{lccccc}
\toprule
\textbf{ignore\_blank} & \textbf{$n$=1} & \textbf{$n$=5} & \textbf{$n$=9} & \textbf{$n$=13} & \textbf{$n$=15} \\
\midrule
Off (1.88) & .799 & .866$^{**}$ & .882$^{**}$ & .883$^{***}$ & .884$^{**}$ \\
On (\textbf{1.12}) & .799 & \textbf{.886} & \textbf{.894} & \textbf{.896} & \textbf{.895} \\
$\Delta$ & -- & +.020 & +.012 & +.013 & +.011 \\
\bottomrule
\end{tabular}
\hfill
\begin{tabular}{lcc}
\toprule
\textbf{Voting} & \textbf{Score} & \textbf{Rank} \\
\midrule
AnswerPriority & \textbf{0.892} & \textbf{1.62} \\
Independent & \underline{0.889} & 1.75 \\
Intersection & 0.890 & 2.62 \\
RefPriority & 0.889 & 4.38 \\
Union & 0.884$^{\dagger}$ & 4.62 \\
\bottomrule
\end{tabular}
\end{table}

Ensemble performance exhibits logarithmic scaling with ensemble size, plateauing at $n$=9--11. The \texttt{ignore\_blank} option consistently improves performance by 1.2 percentage points at $n$=9 (0.882$\rightarrow$0.894) by excluding abstaining runs from voting. \textsc{AnswerPriority} voting achieves the best average rank (1.62) by ensuring reference consistency with the selected answer. Combining optimal configurations achieves \textbf{0.930} on the training set.

\subsection{Error Analysis}
\label{subsec:error_analysis}

We analyzed 2,583 predictions across all LLM sweep runs to identify failure patterns. Of all predictions, 75.2\% were fully correct. Among the 24.8\% errors, the three dominant categories are: \textbf{unnecessary abstention} (26.8\% of errors), where models output \texttt{is\_blank} despite sufficient evidence; \textbf{reference mismatch} (23.6\%), where answers are correct but citations are wrong; and \textbf{value selection errors} (22.2\%), where models select incorrect but contextually plausible values. Rounding/calculation errors account for 13.3\%, type mismatches for 12.2\%, and unit conversion errors for only 1.6\%, indicating that LLMs handle unit consistency well when the correct value is retrieved. Detailed categorization criteria and analysis are provided in Appendix~\ref{app:errors}.

\section{Results}
\label{sec:results}

We evaluate KohakuRAG on the WattBot 2025 Challenge test set, which consists of 282 questions split into public (66\%) and private (34\%) partitions. KohakuRAG achieves \textbf{first place} on the final leaderboard with a private score of \textbf{0.861}. Table~\ref{tab:leaderboard} summarizes our submitted configurations and compares public vs.\ private score dynamics across top teams.

\begin{table}[b]
\caption{WattBot 2025 test set results and leaderboard comparison. \textbf{Left}: Our submitted configurations (all use Jina v4 with hierarchical indexing). \textbf{Right}: Public vs.\ private score dynamics across top teams. Only KohakuRAG maintained \#1 on both partitions.}
\label{tab:leaderboard}
\centering
\small
\begin{tabular}{llccc}
\toprule
\textbf{Model} & \textbf{Config} & $n$ & \textbf{Pub} & \textbf{Priv} \\
\midrule
GPT-oss-120B & No img/BM25 & 5 & .884 & .852 \\
GPT-oss-120B & +BM25 & 7 & .862 & .857 \\
Gemini-3-pro & +Images & 1 & \textbf{.901} & .839 \\
\midrule
\multicolumn{2}{l}{Top-5 mixed ensemble} & -- & .902 & \textbf{.861} \\
\bottomrule
\end{tabular}
\hfill
\begin{tabular}{lccc}
\toprule
\textbf{Team} & \textbf{Public} & \textbf{Private} & $\Delta$ \\
\midrule
\textbf{Ours} & .902 (\#1) & .861 (\#1) & $-$.041 \\
Private \#2 & .803 (\#21) & .858 (\#2) & $+$.055 \\
Private \#3 & .807 (\#20) & .851 (\#3) & $+$.044 \\
\midrule
Public \#2 & .886 (\#2) & .840 (\#4) & $-$.046 \\
Public \#3 & .868 (\#3) & .831 (\#9) & $-$.037 \\
\bottomrule
\end{tabular}
\end{table}

\paragraph{Robustness of Ensemble Methods.}
Ensemble strategies provide substantially more robust predictions across evaluation partitions. The GPT-oss-120B 7-ensemble achieves the most stable performance (0.862 public, 0.857 private, a gap of only 0.5\%). In contrast, single-run Gemini-3-pro exhibits high variance: while achieving the highest public score (0.901), its private score drops to 0.839 ($-$6.2\%). This highlights a trade-off between peak performance and consistency, as single models may overfit to public partition characteristics, whereas ensembles generalize more reliably.

\paragraph{Leaderboard Dynamics.}
The public-to-private transition reveals instructive patterns. Teams ranking \#21 and \#20 publicly achieved \#2 and \#3 privately (+0.055, +0.044), suggesting their methods generalized better despite lower public scores. Conversely, Public \#2 and \#3 fell to Private \#4 and \#9 ($-$0.046, $-$0.037). KohakuRAG was the only submission to maintain first place across both partitions, validating our emphasis on ensemble diversity and hierarchical retrieval robustness.

\paragraph{Single-Model vs. Ensemble Performance.}
Gemini-3-pro with direct image input achieves the highest public score (0.901) in a single inference run, surpassing GPT-oss-120B ensembles. This demonstrates the value of native visual reasoning for documents containing figures, tables, and diagrams, as the model can directly interpret visual elements rather than relying on caption intermediation. However, the 6.2\% degradation on the private partition suggests that single-model predictions are more susceptible to distribution shift between question subsets.

\paragraph{Hybrid Retrieval Impact.}
Adding BM25 sparse retrieval to the GPT-oss-120B ensemble improves private score stability. The 7-ensemble with BM25 achieves 0.857 on the private partition, outperforming the 5-ensemble without BM25 (0.852). This validates the complementary benefits of lexical matching for queries containing specific identifiers or technical terms that may not be captured by dense semantic similarity alone.

\paragraph{Cross-Model Ensemble.}
Our winning submission combines predictions from top-5 configurations spanning multiple model families (Gemini-3-pro, GPT-oss-120B). This cross-model ensemble achieves 0.861 on the private partition, suggesting that different model families capture complementary aspects of the document corpus. The ensemble aggregates Gemini-3-pro's visual reasoning capabilities with GPT-oss-120B's text comprehension strengths, achieving robust performance across diverse question types while mitigating individual model biases.

\section{Conclusion}
\label{sec:conclusion}

We presented KohakuRAG, a hierarchical RAG framework that addresses three limitations of standard RAG for document-grounded question answering: structure loss from flat chunking (via four-level tree indexing with bottom-up embedding propagation), query--document vocabulary mismatch (via LLM-powered multi-query planning with cross-query reranking), and answer instability from single-inference stochasticity (via ensemble voting with abstention-aware blank filtering). Ablation studies show that prompt ordering has outsized impact (+80\% from context-before-question placement), that the retry mechanism effectively addresses unnecessary abstention (the dominant error mode at 26.8\% of failures) with +69\% relative improvement, and that hierarchical dense retrieval alone suffices for citation-heavy tasks (BM25 adds only +3.1pp). Cross-model ensembles provide robustness against distribution shift between evaluation partitions. On the WattBot 2025 Challenge, KohakuRAG achieved first place on both the public and private leaderboards with a final score of 0.861. Notably, we were the only team to maintain the top position across both partitions while other competitive submissions showed score degradation up to $-$0.046.

\paragraph{Limitations.}
Several limitations warrant future investigation. Our rule-based document parsing relies on heuristics for section detection and may fail on irregular layouts. Reference mismatch errors (23.6\% of failures) indicate that LLMs struggle with precise attribution even when answers are correct, often citing survey papers rather than original sources. Ensemble inference incurs linear cost scaling with the number of runs. The retry mechanism improves recall but increases latency for difficult questions. Promising directions include learned document parsing, reference-constrained decoding, and confidence-based adaptive retrieval that predicts when additional context is needed.

\section*{Acknowledgements}
We thank Comfy Org for supporting the LLM inference costs used in this work.

% \newpage
\bibliographystyle{plainnat}
\bibliography{reference}

\clearpage
\appendix
\clearpage

\noindent\rule{\textwidth}{1pt}
\begin{center}
\vspace{3pt}
{\Large Appendix}
\vspace{-3pt}
\end{center}
\noindent\rule{\textwidth}{1pt}

\section*{Table of Contents}

% Redefine how sections appear in TOC
\definecolor{sectionblue}{RGB}{65, 105, 225}
% Format for main sections
\titlecontents{section}
  [1.6em]                                           % No indentation for main sections
  {\addvspace{0.5pc}\color{sectionblue}}          % Above code and blue color
  {\contentslabel[\thecontentslabel]{1.75em}}      % Numbered format
  {\hspace*{-1.5em}}                              % Unnumbered format
  {\hfill\contentspage}                           % Filler and page
  []                                              % Below code

% Format for subsections - with proper indentation
\titlecontents{subsection}
  [4em]                                         % Indentation for subsections
  {\addvspace{0.2pc}\color{sectionblue}}          % Above code and blue color
  {\contentslabel[\thecontentslabel]{2.5em}}      % Numbered format
  {\hspace*{-2.5em}}                              % Unnumbered format
  {\titlerule*[1pc]{.}\contentspage}              % Dots and page number
  []                                              % Below code

\startcontents[appendix]
\printcontents[appendix]{l}{1}{\setcounter{tocdepth}{2}}

\newpage

\section{Implementation Details}
\label{app:implementation}

This appendix provides comprehensive implementation details for KohakuRAG, including document parsing, embedding pipelines, storage, and inference configurations.

\subsection{Indexing Pipeline}

This section describes the document parsing, embedding computation, and vector storage components that transform raw documents into a searchable hierarchical index.

\paragraph{Document Parsing.}
Document parsing employs a rule-based pipeline tailored for technical PDFs and slide decks. For PDF documents, we use PyMuPDF to extract text with position information, then apply heuristics based on font size, spacing, and formatting to identify section boundaries. Slide decks are parsed page-by-page, treating each slide as a section. The parser outputs a JSON tree structure where each node contains:
\begin{itemize}[noitemsep]
    \item \texttt{id}: Hierarchical identifier (e.g., \texttt{doc1:sec2:p3:s4})
    \item \texttt{level}: One of \{\texttt{document}, \texttt{section}, \texttt{paragraph}, \texttt{sentence}\}
    \item \texttt{content}: Raw text content
    \item \texttt{children}: List of child node IDs
    \item \texttt{metadata}: Source file, page number, bounding box (for images)
\end{itemize}

\paragraph{PDF Text Extraction.}
Text extraction uses PyMuPDF's \texttt{get\_text("dict")} method, which returns structured information including bounding boxes, font sizes, and font names. We use the following heuristics:
\begin{itemize}[noitemsep]
    \item \textbf{Section detection}: Text blocks with font size $\geq 1.2\times$ the median font size are treated as potential section headers. Numbered patterns (e.g., ``1.2'', ``A.3'') further confirm section boundaries.
    \item \textbf{Paragraph detection}: Consecutive text blocks with vertical spacing $< 1.5\times$ line height are merged into paragraphs. Larger gaps indicate paragraph boundaries.
    \item \textbf{Sentence segmentation}: We use NLTK's \texttt{sent\_tokenize} with custom rules for handling common abbreviations in technical text (e.g., ``Fig.'', ``et al.'', ``i.e.'').
\end{itemize}

\paragraph{Image Extraction.}
Image extraction uses PyMuPDF's \texttt{get\_images()} to identify embedded figures, which are saved as PNG files. We handle PDF soft masks (SMask) correctly to preserve transparency for diagrams and charts. Captions are generated via Qwen3-VL with a prompt requesting concise descriptions of technical content, data values, and axis labels for charts:

\begin{lstlisting}
Describe this image from a technical document.
Focus on:
1. Data values and numerical information
2. Axis labels and units for charts
3. Key visual elements and their relationships
Keep the description concise (2-3 sentences).
\end{lstlisting}

\paragraph{Embedding Pipeline.}
The embedding pipeline processes nodes bottom-up through the hierarchy. For Jina v3, we use the official API with \texttt{task=retrieval.passage} for indexing and \texttt{task=retrieval.query} for queries. For Jina v4, we use \texttt{jina-clip-v2} which natively supports both text and image inputs.

Algorithm~\ref{alg:embedding} presents the bottom-up embedding propagation procedure.

\begin{algorithm}[H]
\caption{Bottom-Up Embedding Propagation}
\label{alg:embedding}
\begin{algorithmic}[1]
\Require Document tree $T = (V, E)$, encoder $E$
\Ensure Embeddings $\{\mathbf{e}_v\}_{v \in V}$
\Statex
\Function{EmbedTree}{$T, E$}
    \State $\mathcal{L} \gets \{v \in V : \mathcal{C}(v) = \emptyset\}$ \Comment{Collect leaf nodes}
    \For{$v \in \mathcal{L}$} \Comment{Embed leaves (batched)}
        \State $\mathbf{e}_v \gets E(t_v)$
    \EndFor
    \Statex
    \For{level $\in$ [paragraph, section, document]} \Comment{Bottom-up}
        \For{$v$ at current level}
            \State $\mathbf{e}_v \gets \frac{\sum_{c \in \mathcal{C}(v)} |t_c| \cdot \mathbf{e}_c}{\sum_{c \in \mathcal{C}(v)} |t_c|}$ \Comment{Weighted average}
        \EndFor
    \EndFor
    \State \Return $\{\mathbf{e}_v\}_{v \in V}$
\EndFunction
\end{algorithmic}
\end{algorithm}

\paragraph{Embedding Dimensions.}
Jina v4 supports Matryoshka representations with configurable output dimensions (128--2048). We use 512 dimensions as the default, balancing retrieval quality with storage efficiency. Matryoshka truncation is applied post-embedding by slicing vectors to the target dimension.

\paragraph{Vector Storage.}
We use sqlite-vec (v0.1.1), a SQLite extension providing approximate nearest neighbor search via IVF-PQ indexing. The schema stores:
\begin{itemize}[noitemsep]
    \item \texttt{nodes} table: node metadata (id, level, content, parent\_id)
    \item \texttt{embeddings} virtual table: vector index with node\_id foreign key
\end{itemize}

KohakuVault wraps sqlite-vec with a Python interface supporting batch insert, filtered search (by level), and automatic index rebuilding when the corpus changes.

\subsection{Retrieval Pipeline}

This section describes the query planning, cross-query reranking, and optional BM25 hybrid retrieval components that transform user questions into ranked document passages.

\paragraph{Query Planning.}
The query planner uses a structured prompt requesting $n$ diverse queries:
\begin{lstlisting}
Given the question: {question}
Generate {n} search queries that:
1. Rephrase using alternative terminology
2. Expand abbreviations (e.g., PUE -> Power Usage Effectiveness)
3. Add likely keywords from technical documents
4. Decompose into sub-questions if compound
Output as JSON array of strings.
\end{lstlisting}

The planner LLM (same as generator) returns queries parsed via JSON; malformed outputs fall back to the original question only.

\paragraph{Example Query Expansion.}
For the question ``What is the PUE of Google's data centers?'', the planner generates:
\begin{enumerate}[noitemsep]
    \item ``Google data center Power Usage Effectiveness''
    \item ``Google infrastructure energy efficiency ratio''
    \item ``PUE metric Google cloud computing facilities''
    \item ``Google sustainability report data center efficiency''
\end{enumerate}

\paragraph{Cross-Query Reranking.}
Algorithm~\ref{alg:rerank} presents the cross-query reranking procedure that aggregates results from multiple planned queries.

\begin{algorithm}[H]
\caption{Cross-Query Reranking}
\label{alg:rerank}
\begin{algorithmic}[1]
\Require Query results $\{\mathcal{R}_i\}_{i=1}^{n}$ where $\mathcal{R}_i = \{(v, s_{iv})\}$, strategy $\in$ \{\textsc{Freq}, \textsc{Score}, \textsc{Combined}\}
\Ensure Ranked list of nodes
\Statex
\Function{Rerank}{$\{\mathcal{R}_i\}$, strategy}
    \State $\mathcal{V} \gets \bigcup_i \{v : (v, s) \in \mathcal{R}_i\}$ \Comment{Unique nodes}
    \For{$v \in \mathcal{V}$}
        \State $f_v \gets |\{i : v \in \mathcal{R}_i\}|$ \Comment{Frequency}
        \State $s_v \gets \sum_{i : v \in \mathcal{R}_i} s_{iv}$ \Comment{Total score}
    \EndFor
    \Statex
    \If{strategy = \textsc{Frequency}}
        \State \Return $\textsc{SortBy}(\mathcal{V}, (f_v, s_v), \text{desc})$
    \ElsIf{strategy = \textsc{Score}}
        \State \Return $\textsc{SortBy}(\mathcal{V}, s_v, \text{desc})$
    \ElsIf{strategy = \textsc{Combined}}
        \State $\hat{f}_v \gets (f_v - \min f) / (\max f - \min f)$ \Comment{Normalize}
        \State $\hat{s}_v \gets (s_v - \min s) / (\max s - \min s)$
        \State \Return $\textsc{SortBy}(\mathcal{V}, 0.5 \cdot \hat{f}_v + 0.5 \cdot \hat{s}_v, \text{desc})$
    \EndIf
\EndFunction
\end{algorithmic}
\end{algorithm}

\paragraph{BM25 Indexing.}
BM25 indexing uses rank\_bm25 library with default parameters ($k_1=1.5$, $b=0.75$). We index paragraph-level content only (not sentences) to balance precision with recall. Tokenization uses NLTK word\_tokenize with lowercasing and stopword removal.

\paragraph{Hybrid Retrieval.}
At query time, BM25 scores are computed for all paragraphs, and top-$k_{\text{bm25}}$ results are appended to dense retrieval results without score normalization. This simple concatenation strategy avoids the need to calibrate score distributions between dense and sparse retrievers.

\paragraph{Tree Deduplication.}
Because retrieval targets both sentence and paragraph nodes, multi-query retrieval can return a parent node alongside one or more of its descendants, e.g., paragraph \texttt{doc:sec1:p2} and sentence \texttt{doc:sec1:p2:s3}. Standard node-ID deduplication only removes exact duplicates and does not detect this hierarchical overlap; the child's text is already contained within the parent, wasting context window capacity. Tree deduplication addresses this by removing any node whose ancestor (determined by node-ID prefix matching) is also present in the result set. This step is applied after reranking but before top-$k_{\text{final}}$ truncation, so that the freed slots can be filled by other unique content.

Algorithm~\ref{alg:tree_dedup} presents the tree deduplication procedure. The algorithm first collects all unique node IDs from the expanded snippet set, then identifies nodes whose ancestor is also present by checking node-ID prefix relationships. Finally, it removes these subsumed nodes and deduplicates by node ID (keeping the first occurrence) in a single pass.

\begin{algorithm}[H]
\caption{Tree Deduplication of Context Snippets}
\label{alg:tree_dedup}
\begin{algorithmic}[1]
\Require Expanded snippet set $\mathcal{S} = \{(v_i, t_i, s_i)\}$ where $v_i$ is node ID, $t_i$ is text, $s_i$ is score
\Ensure Deduplicated snippets $\mathcal{S}'$ with no ancestor--descendant overlap
\Statex
\Function{TreeDedup}{$\mathcal{S}$}
    \State $\mathcal{V} \gets \{v_i : (v_i, t_i, s_i) \in \mathcal{S}\}$ \Comment{Collect unique node IDs}
    \State $\mathcal{D} \gets \emptyset$ \Comment{IDs to discard}
    \For{$v \in \mathcal{V}$}
        \For{$u \in \mathcal{V} \setminus \{v\}$}
            \If{$v$ starts with $u$\texttt{:}} \Comment{$u$ is ancestor of $v$}
                \State $\mathcal{D} \gets \mathcal{D} \cup \{v\}$; \textbf{break}
            \EndIf
        \EndFor
    \EndFor
    \Statex
    \State $\mathcal{S}' \gets [\ ]$; $\textit{seen} \gets \emptyset$
    \For{$(v_i, t_i, s_i) \in \mathcal{S}$} \Comment{Preserve original order}
        \If{$v_i \notin \mathcal{D}$ \textbf{and} $v_i \notin \textit{seen}$}
            \State append $(v_i, t_i, s_i)$ to $\mathcal{S}'$; $\textit{seen} \gets \textit{seen} \cup \{v_i\}$
        \EndIf
    \EndFor
    \State \Return $\mathcal{S}'$
\EndFunction
\end{algorithmic}
\end{algorithm}

We evaluated the context reduction on the WattBot training set (41 questions) using Jina v4 embeddings with 4 planner queries $\times$ 16 retrievals per query and top-$k_{\text{final}}$=32. Table~\ref{tab:tree_dedup} reports aggregate statistics. Without deduplication, the 4$\times$16=64 raw matches are truncated to 32, yielding 148K characters of context with substantial redundancy from nodes retrieved by multiple queries. Node-ID dedup removes these cross-query duplicates, reducing context by 35.0\%. Tree dedup further removes hierarchical overlap, achieving a cumulative 51.4\% reduction (from 148.1K to 72.0K characters) without discarding any unique information, since the parent node already subsumes all descendant text. Although tree deduplication was not enabled in our competition submissions, it provides a practical mechanism for reducing context length in deployment scenarios where LLM context windows or API costs are constrained.

\begin{table}[H]
\centering
\caption{Context statistics under different deduplication modes (mean $\pm$ std over 41 training questions, 4 queries $\times$ 16 retrievals each). Node-ID dedup removes cross-query duplicates ($-$35.0\%); tree dedup additionally removes hierarchical overlap ($-$51.4\% cumulative).}
\label{tab:tree_dedup}
\small
\begin{tabular}{lccc}
\toprule
\textbf{Mode} & \textbf{Matches} & \textbf{Snippets} & \textbf{Context (chars)} \\
\midrule
No dedup       & 32.0 $\pm$ 0.0 & 64.0 $\pm$ 0.0 & 148,125 $\pm$ 34,415 \\
Node-ID dedup  & 20.9 $\pm$ 3.9 & 41.8 $\pm$ 7.8 &  96,332 $\pm$ 24,577 \\
Tree dedup     & 15.7 $\pm$ 4.4 & 31.4 $\pm$ 8.9 &  71,956 $\pm$ 23,449 \\
\bottomrule
\end{tabular}
\end{table}

\subsection{Answer Generation Pipeline}

This section describes the prompt construction, structured output format, and ensemble voting mechanism that transform retrieved context into final answers with citations.

\paragraph{Prompt Construction.}
Retrieved nodes are formatted as:
\begin{lstlisting}
[ref_id={doc_id}] {node_content}
\end{lstlisting}

The full prompt structure (with reordering):
\begin{lstlisting}
## Referenced Documents
{formatted_nodes}

## Referenced Media
{image_captions}

## Question
{question}

## Instructions
Answer the question using only the provided references.
If the answer is numeric, extract the exact value with units.
If evidence is insufficient, set is_blank=true.

Output JSON with fields:
- answer: Natural language answer
- answer_value: Extracted numeric or categorical value
- ref_id: List of source document IDs
- explanation: Brief reasoning (optional)
- is_blank: true if cannot answer from context
\end{lstlisting}

\paragraph{Context Length Management.}
We limit total context to 8,000 tokens to avoid degradation from overly long contexts. When retrieved content exceeds this limit, we truncate lower-ranked nodes while preserving the hierarchical context of top-ranked nodes.

\paragraph{Ensemble Voting.}
Ensemble voting aggregates $m$ independent runs. For answer voting, we normalize answers (lowercase, strip whitespace, parse numerics) before counting. Algorithm~\ref{alg:ensemble} presents the complete voting procedure.

\begin{algorithm}[H]
\caption{Ensemble Voting with Blank Handling}
\label{alg:ensemble}
\begin{algorithmic}[1]
\Require Predictions $\mathcal{A} = \{(a^{(j)}, R^{(j)}, b^{(j)})\}_{j=1}^{m}$, \texttt{ignore\_blank}, \texttt{vote\_mode}
\Ensure Final prediction $(a, R, b)$
\Statex
\Function{EnsembleVote}{$\mathcal{A}$, \texttt{ignore\_blank}, \texttt{vote\_mode}}
    \If{\texttt{ignore\_blank} \textbf{and} $\exists j: b^{(j)} = 0$}
        \State $\mathcal{A} \gets \{(a^{(j)}, R^{(j)}, b^{(j)}) \in \mathcal{A} : b^{(j)} = 0\}$ \Comment{Filter blanks}
    \EndIf
    \Statex
    \If{$\forall j: b^{(j)} = 1$} \Comment{All abstained}
        \State \Return $(\texttt{null}, \emptyset, 1)$
    \EndIf
    \Statex
    \State $\mathcal{G} \gets \textsc{GroupBy}(\mathcal{A}, \textsc{Normalize}(a))$ \Comment{Group by answer}
    \State $a \gets \arg\max_{g \in \mathcal{G}} |g|$ \Comment{Plurality vote}
    \State $\mathcal{A}_a \gets \{(a', R', b') \in \mathcal{A} : \textsc{Normalize}(a') = a\}$
    \Statex
    \If{\texttt{vote\_mode} = \textsc{AnswerPriority}}
        \State $R \gets \textsc{MajorityVote}(\{R' : (a', R', b') \in \mathcal{A}_a\})$
    \ElsIf{\texttt{vote\_mode} = \textsc{Union}}
        \State $R \gets \bigcup_{(a', R', b') \in \mathcal{A}_a} R'$
    \ElsIf{\texttt{vote\_mode} = \textsc{Intersection}}
        \State $R \gets \bigcap_{(a', R', b') \in \mathcal{A}_a} R'$
    \EndIf
    \State \Return $(a, R, 0)$
\EndFunction
\end{algorithmic}
\end{algorithm}

\paragraph{Reference Aggregation Strategies.}
We support four strategies for combining reference sets across ensemble runs:
\begin{itemize}[noitemsep]
    \item \textsc{Independent}: Majority vote on references separately from answer
    \item \textsc{AnswerPriority}: Collect refs only from runs matching winning answer
    \item \textsc{Union}: Union of refs from runs matching winning answer
    \item \textsc{Intersection}: Intersection of refs from runs matching winning answer
\end{itemize}

\paragraph{Numeric Answer Normalization.}
For numeric answers, we parse values and compare within a tolerance:
\begin{itemize}[noitemsep]
    \item Parse scientific notation (e.g., ``1.5e6'' $\rightarrow$ 1,500,000)
    \item Handle common units (e.g., ``1.5 MW'' $\rightarrow$ 1,500,000 W)
    \item Group answers within 0.1\% relative difference as equivalent
\end{itemize}

\subsection{Hyperparameter Configuration}

Table~\ref{tab:hyperparams} summarizes the hyperparameter search space and defaults used throughout our experiments.

\begin{table}[H]
\centering
\caption{Hyperparameter search space and defaults.}
\label{tab:hyperparams}
\small
\begin{tabular}{llll}
\toprule
\textbf{Parameter} & \textbf{Values Tested} & \textbf{Default} & \textbf{Notes} \\
\midrule
top-$k$ (per query) & 4, 8, 16 & 16 & Higher = more coverage \\
top-$k_{\text{final}}$ & 16, 32, None & 32 & Post-rerank truncation \\
Planner queries & 2, 4, 6 & 4 & More = better coverage \\
Max retries & 0, 1, 2 & 0 & On abstention \\
BM25 top-$k$ & 0, 2, 4, 8 & 0 & 0 = dense only \\
Ensemble size & 1, 3, 5, 7, 9, 11, 13, 15 & 1 & Odd numbers only \\
Embedding dim (v4) & 128, 256, 512, 1024, 2048 & 512 & Matryoshka \\
Temperature & 0.0, 0.3, 0.7, 1.0 & 0.7 & For ensemble \\
\bottomrule
\end{tabular}
\end{table}
\section{Complete Sweep Results}
\label{app:sweeps}

This appendix presents comprehensive results from our hyperparameter sweeps, providing detailed empirical evidence supporting the design choices described in Section~\ref{sec:experiment}. Understanding these trade-offs is valuable for practitioners adapting KohakuRAG to different domains or computational constraints.

\subsection{Experimental Setup}

All experiments were conducted on the WattBot 2025 training set (176 questions) using a consistent evaluation protocol. Each configuration reports mean score over 3 random seeds unless otherwise noted. Ensemble experiments use 32 stratified subsamples to estimate variance, with the same base predictions used across all ensemble sizes to enable fair comparison.

We organize results by component category: embedding configuration, image handling, retrieval parameters, context enhancement, LLM selection, and ensemble inference. For each category, we present both visualizations (figures) and detailed numerical results (tables) to support both quick intuition-building and precise comparison.

\subsection{Embedding and Multimodal Configuration}

\begin{figure}[H]
\centering
\begin{minipage}[t]{0.48\textwidth}
\centering
\includegraphics[width=\textwidth]{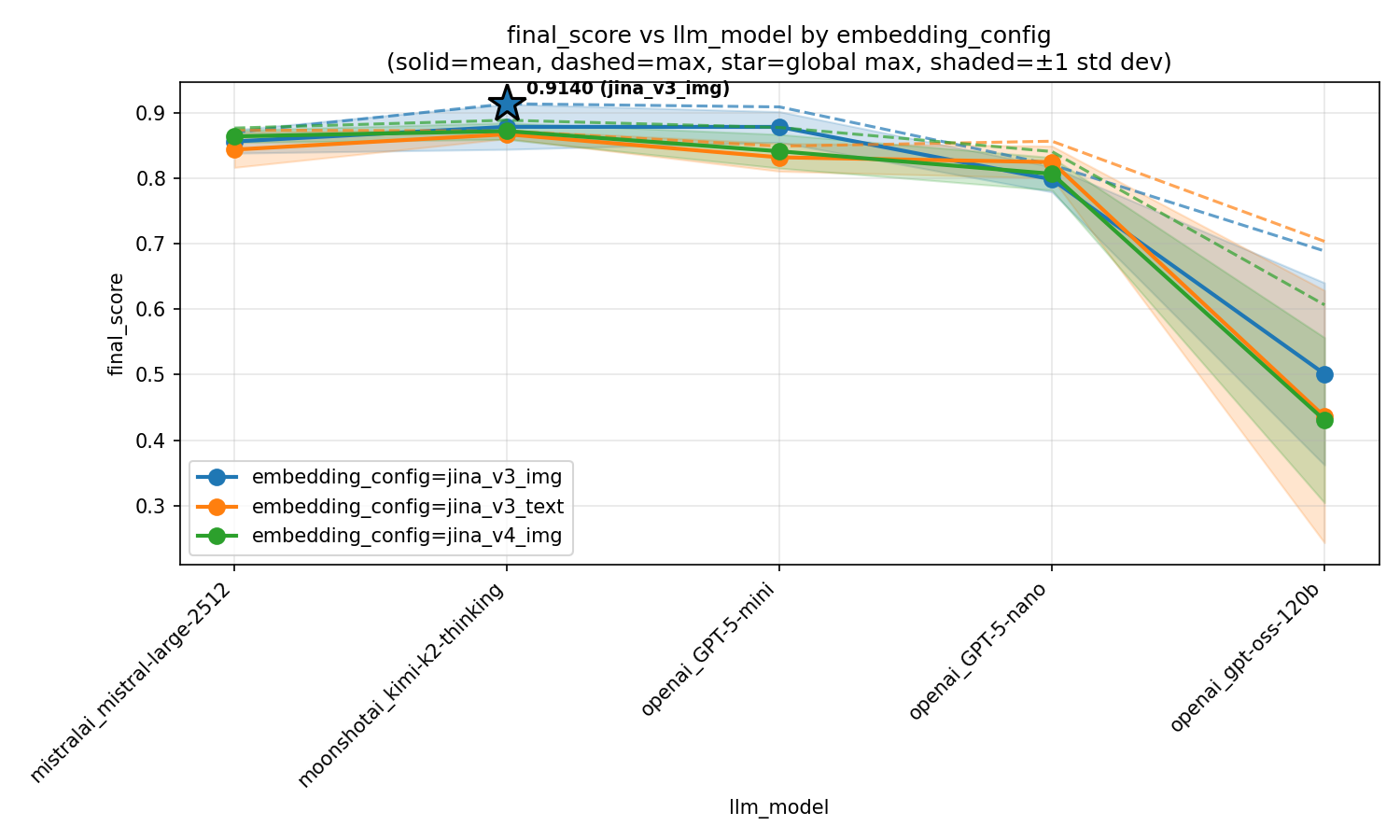}
\caption{LLM $\times$ Embedding configuration.}
\label{fig:llm_embedding}
\end{minipage}
\hfill
\begin{minipage}[t]{0.48\textwidth}
\centering
\includegraphics[width=\textwidth]{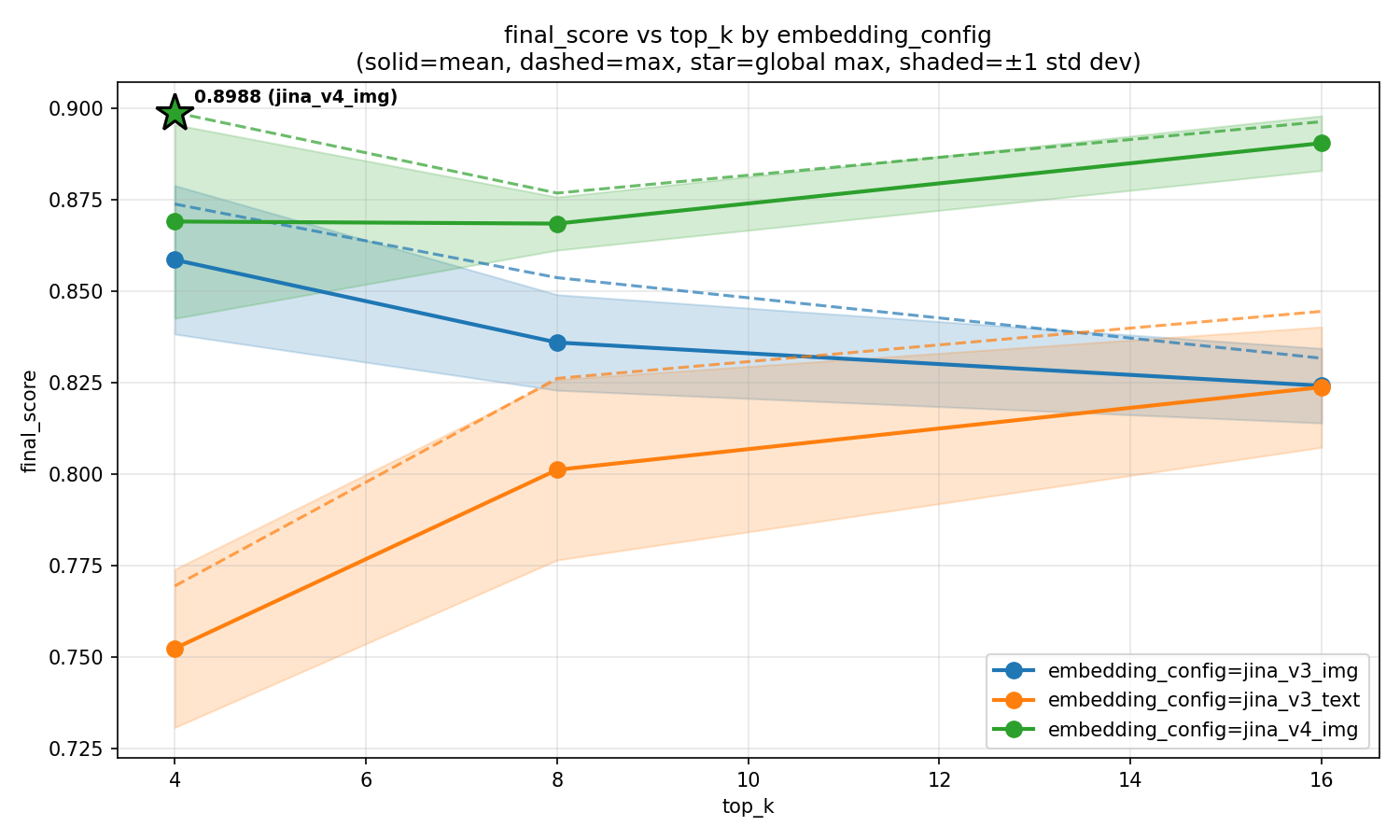}
\caption{Top-$k$ $\times$ Embedding configuration.}
\label{fig:topk_embedding}
\end{minipage}
\end{figure}

Figure~\ref{fig:llm_embedding} shows the interaction between LLM choice and embedding configuration. Jina v4 with native image embeddings consistently outperforms text-only Jina v3 across most LLMs, with the exception of GPT-oss-120B which shows degraded performance across all configurations (likely due to instruction-following limitations). Figure~\ref{fig:topk_embedding} demonstrates that the advantage of multimodal embeddings is most pronounced at lower retrieval depths ($k$=4), where precise retrieval is critical.

\begin{table}[H]
\centering
\caption{LLM $\times$ Embedding configuration (final score).}
\label{tab:llm_embedding_full}
\small
\begin{tabular}{lccc}
\toprule
\textbf{Model} & \textbf{Jina v3 text} & \textbf{Jina v3 +img} & \textbf{Jina v4 +img} \\
\midrule
GPT-5-nano & 0.825 & 0.798 & 0.807 \\
GPT-5-mini & 0.832 & 0.879 & 0.841 \\
GPT-oss-120B & 0.436 & 0.501 & 0.431 \\
Mistral-large-2512 & 0.844 & 0.856 & 0.864 \\
Kimi-k2-thinking & 0.867 & 0.879 & 0.872 \\
\bottomrule
\end{tabular}
\end{table}

\subsection{Image Retrieval Analysis}

\begin{figure}[H]
\centering
\includegraphics[width=0.6\textwidth]{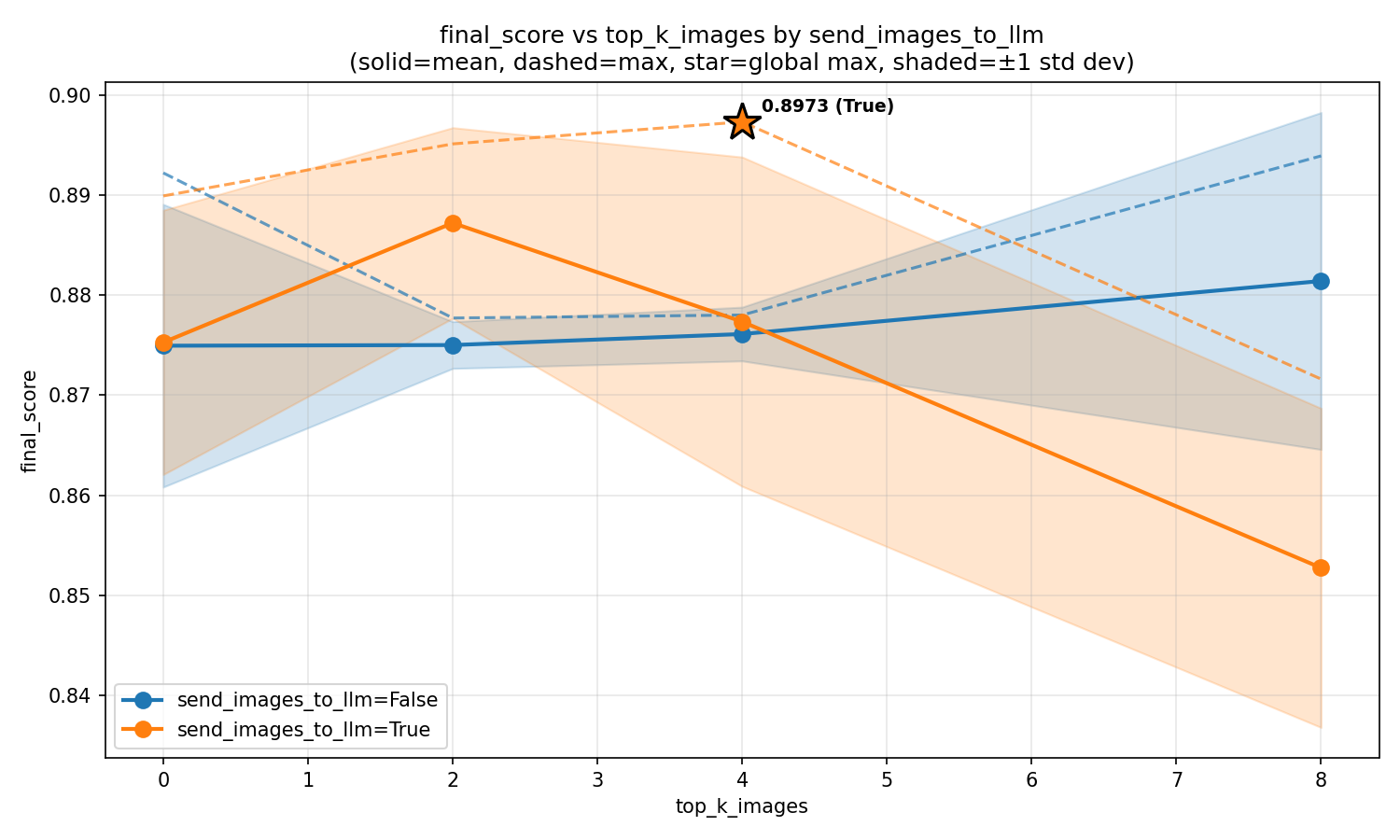}
\caption{Send images to LLM $\times$ Image top-$k$ configuration. Sending 2 images provides optimal performance.}
\label{fig:send_images}
\end{figure}

We evaluate whether sending actual images (rather than just captions) to vision-capable LLMs improves answer quality. As shown in Figure~\ref{fig:send_images} and Table~\ref{tab:image_retrieval}, sending images provides modest improvement (+1.2\% at img $k$=2), but degrades with too many images ($k$=8), likely due to context dilution. This suggests a ``sweet spot'' exists for multimodal context: enough images to provide visual evidence, but not so many that they overwhelm textual reasoning.

\begin{table}[H]
\centering
\caption{Image retrieval configuration with Grok-4.1-fast (final score).}
\label{tab:image_retrieval}
\small
\begin{tabular}{lcccc}
\toprule
\textbf{Send to LLM} & \textbf{img $k$=0} & \textbf{img $k$=2} & \textbf{img $k$=4} & \textbf{img $k$=8} \\
\midrule
False (caption only) & 0.875 & 0.875 & 0.876 & 0.881 \\
True (with images) & 0.875 & \textbf{0.887} & 0.877 & 0.853 \\
\bottomrule
\end{tabular}
\end{table}

\subsection{Retrieval Configuration}

\begin{figure}[H]
\centering
\begin{minipage}[t]{0.48\textwidth}
\centering
\includegraphics[width=\textwidth]{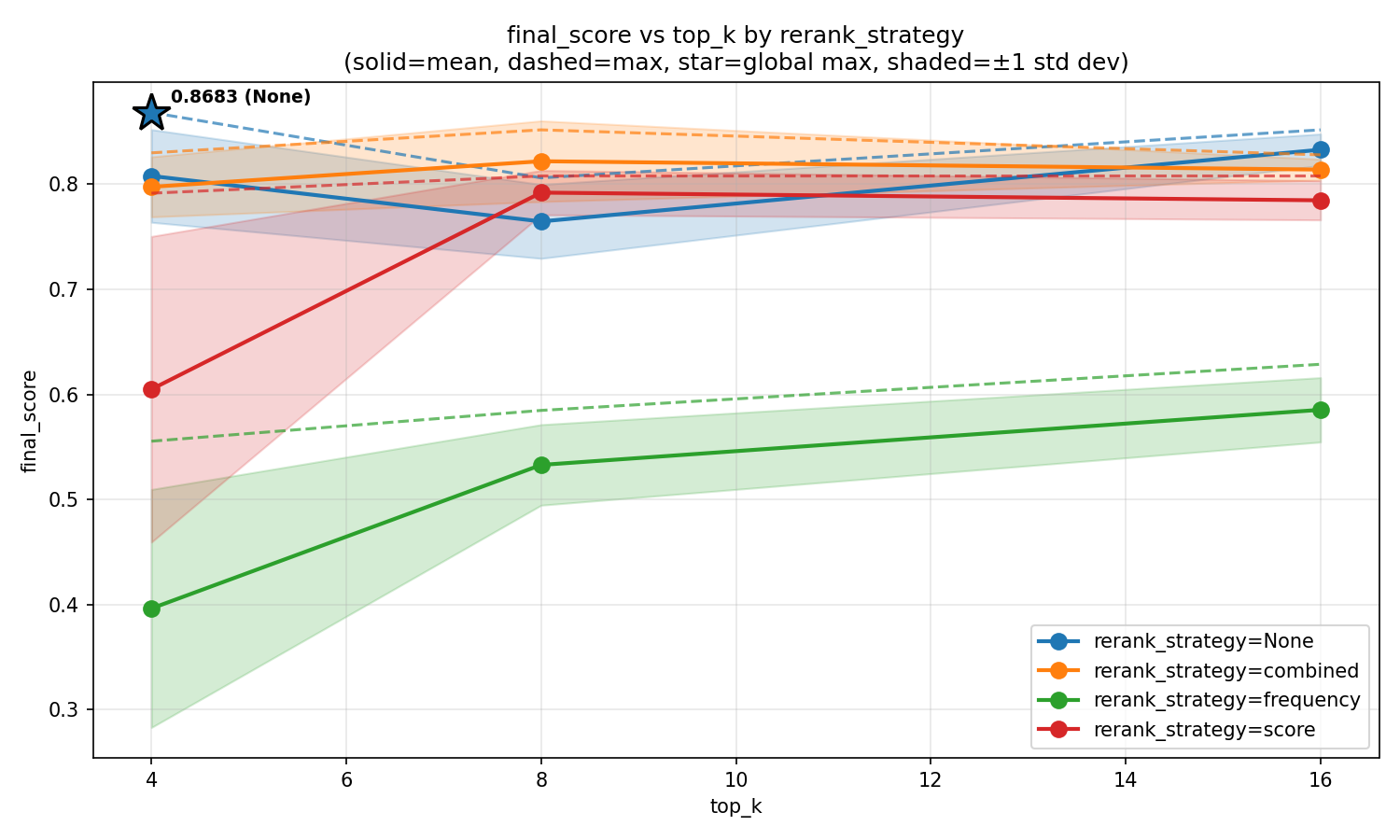}
\caption{Top-$k$ $\times$ Reranking strategy.}
\label{fig:topk_rerank}
\end{minipage}
\hfill
\begin{minipage}[t]{0.48\textwidth}
\centering
\includegraphics[width=\textwidth]{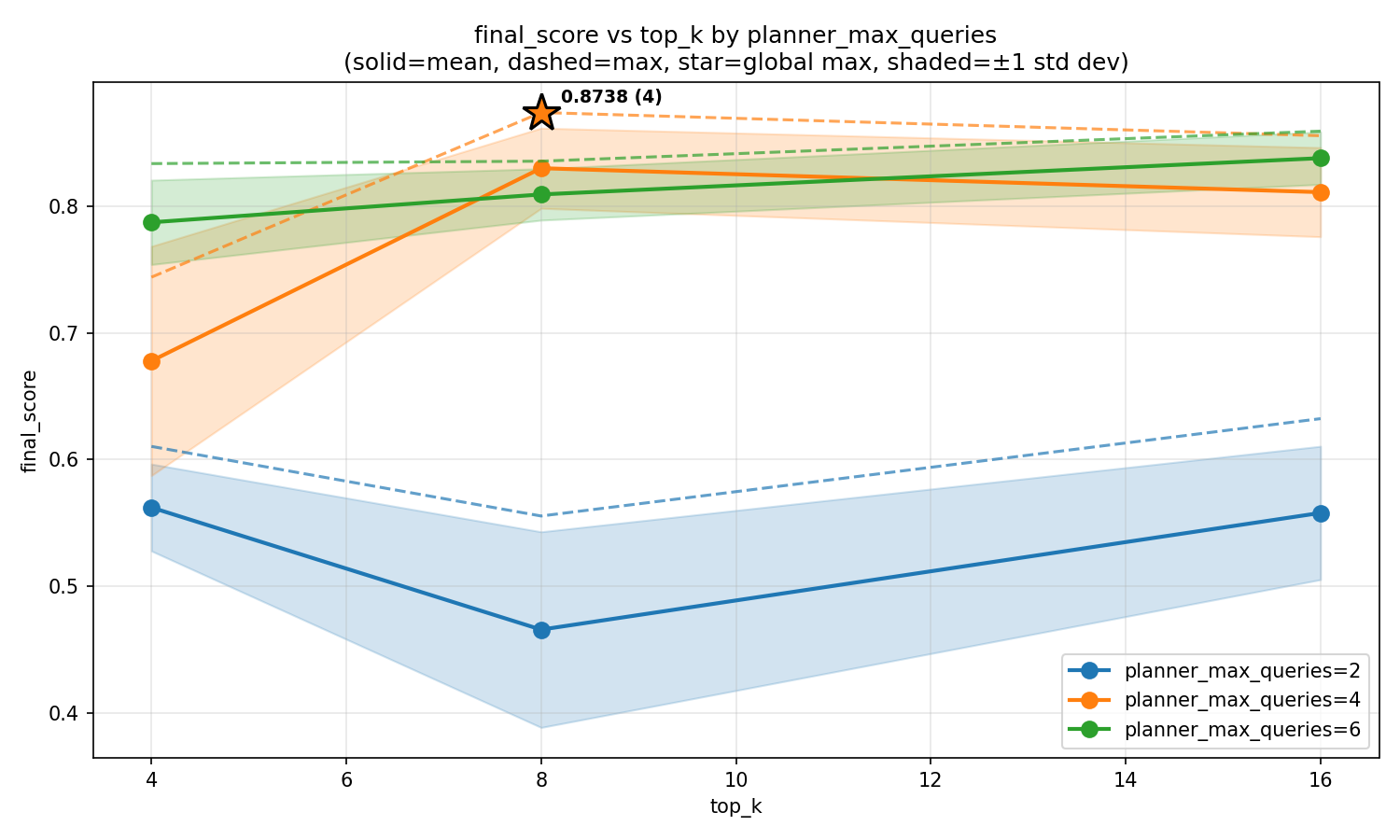}
\caption{Planner queries $\times$ Top-$k$.}
\label{fig:planner_topk}
\end{minipage}
\end{figure}

\begin{figure}[H]
\centering
\begin{minipage}[t]{0.48\textwidth}
\centering
\includegraphics[width=\textwidth]{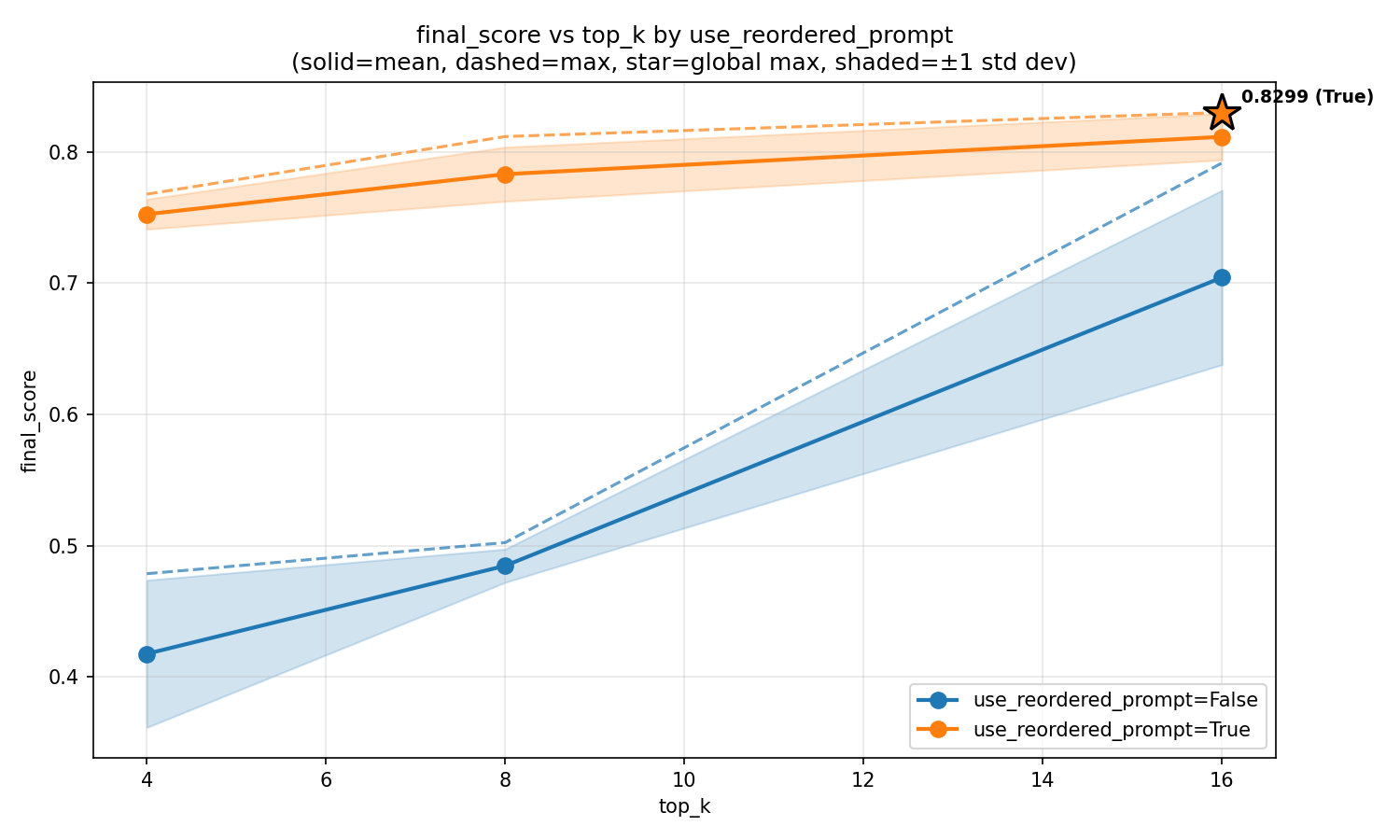}
\caption{Top-$k$ $\times$ Prompt reordering.}
\label{fig:topk_reorder}
\end{minipage}
\hfill
\begin{minipage}[t]{0.48\textwidth}
\centering
\includegraphics[width=\textwidth]{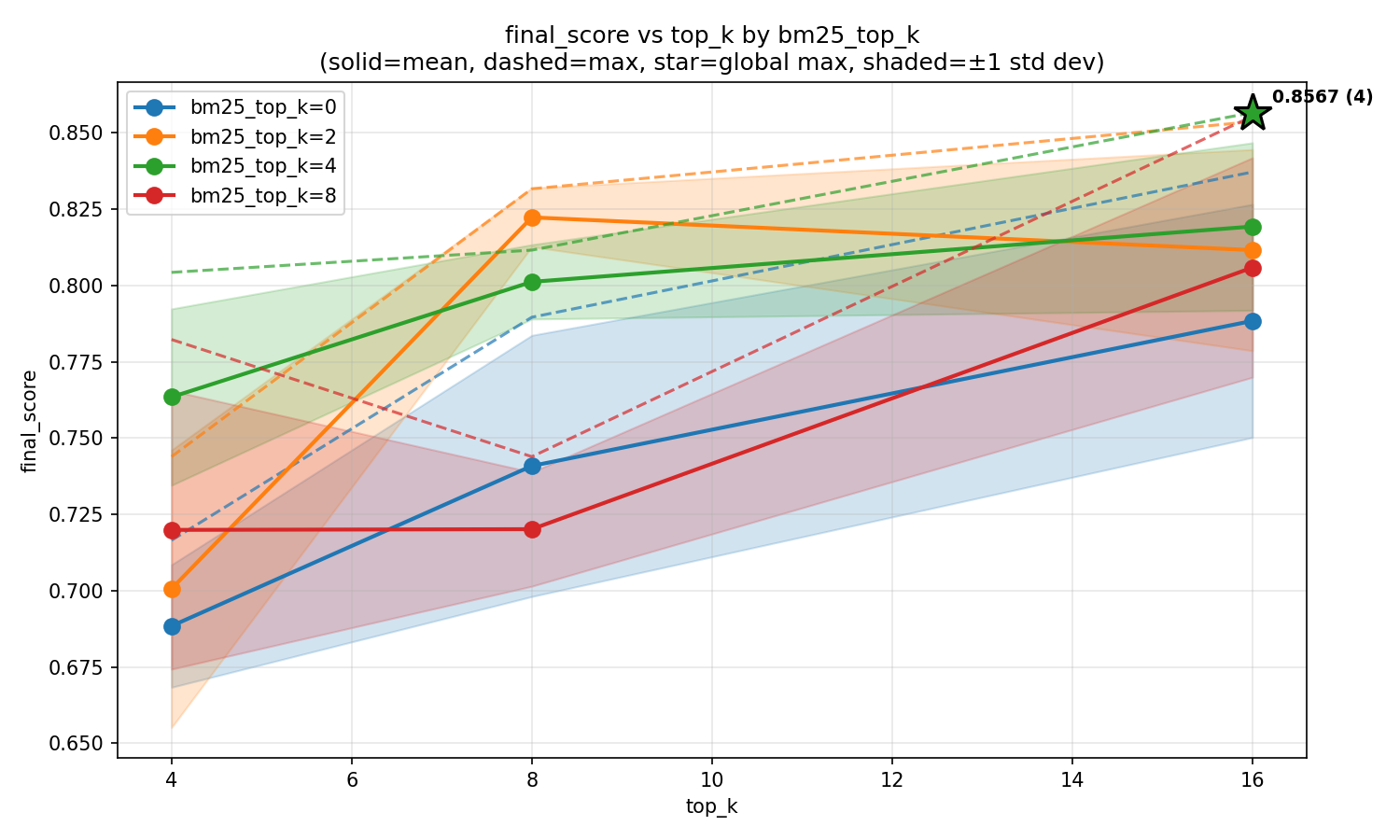}
\caption{BM25 top-$k$ $\times$ Dense top-$k$.}
\label{fig:bm25_topk}
\end{minipage}
\end{figure}

Figures~\ref{fig:topk_rerank}--\ref{fig:bm25_topk} present ablations on retrieval parameters. Key findings:

\paragraph{Reranking Strategy (Figure~\ref{fig:topk_rerank}).} The \textsc{Combined} strategy (frequency $+$ score) provides the most consistent results across retrieval depths. Pure frequency-based reranking severely underperforms, indicating that while document frequency provides useful signal, retrieval scores remain important.

\paragraph{Query Planning (Figure~\ref{fig:planner_topk}).} Increasing planner queries from 2 to 6 improves performance substantially at all retrieval depths, with the largest gains at $k$=4 (+40\%). More queries help capture diverse aspects of complex questions.

\paragraph{Prompt Reordering (Figure~\ref{fig:topk_reorder}).} Placing retrieved context before the question (C$\rightarrow$Q order) dramatically outperforms the standard order, confirming the ``lost in the middle'' phenomenon. The effect is largest at lower $k$ values where every piece of context matters.

\paragraph{BM25 Hybrid Search (Figure~\ref{fig:bm25_topk}).} Adding BM25 provides consistent improvements over dense-only retrieval, with optimal performance at BM25 $k$=4. Higher BM25 $k$ values dilute the precision advantage.

\subsection{Context Enhancement}

\begin{figure}[H]
\centering
\begin{minipage}[t]{0.48\textwidth}
\centering
\includegraphics[width=\textwidth]{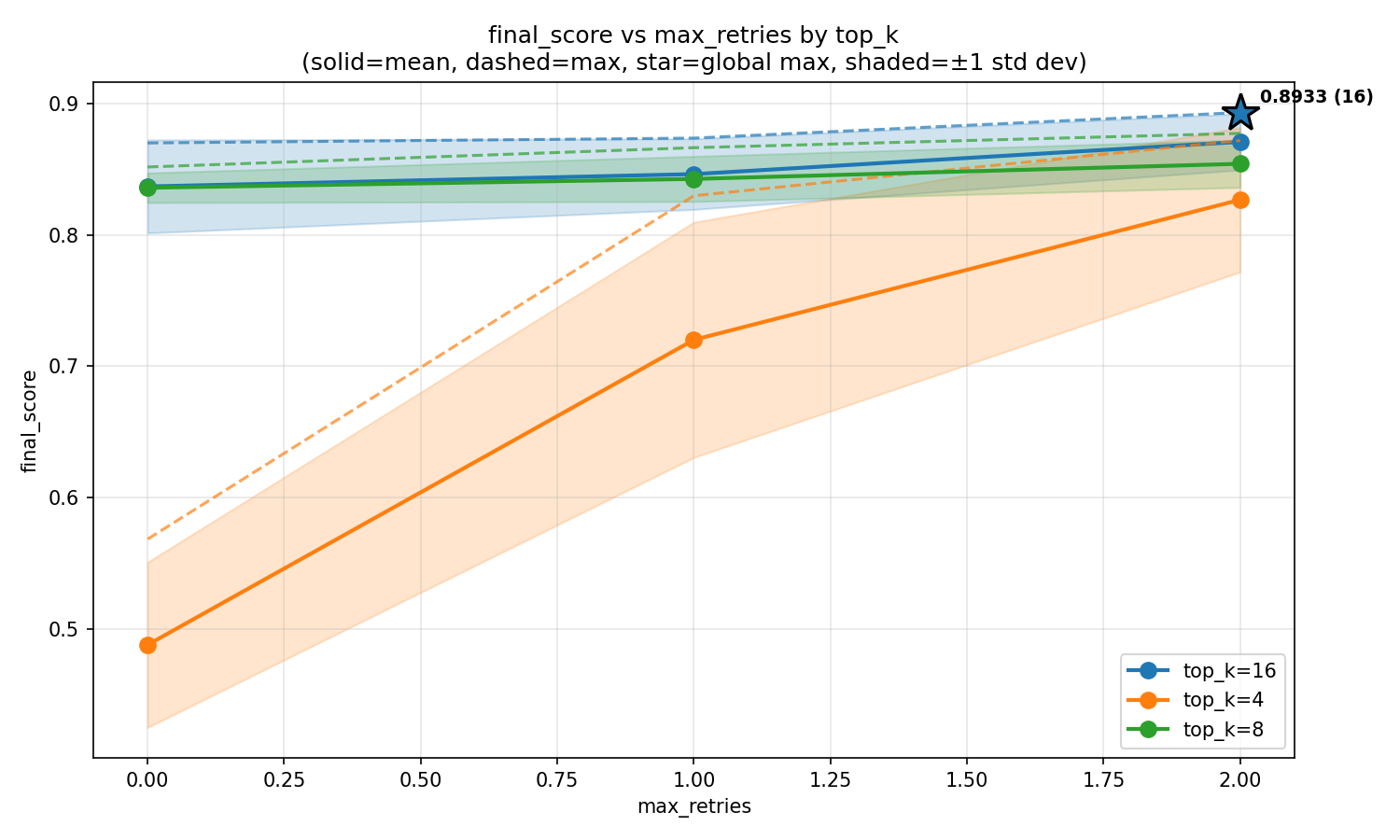}
\caption{Top-$k$ $\times$ Max retries.}
\label{fig:topk_retries}
\end{minipage}
\hfill
\begin{minipage}[t]{0.48\textwidth}
\centering
\includegraphics[width=\textwidth]{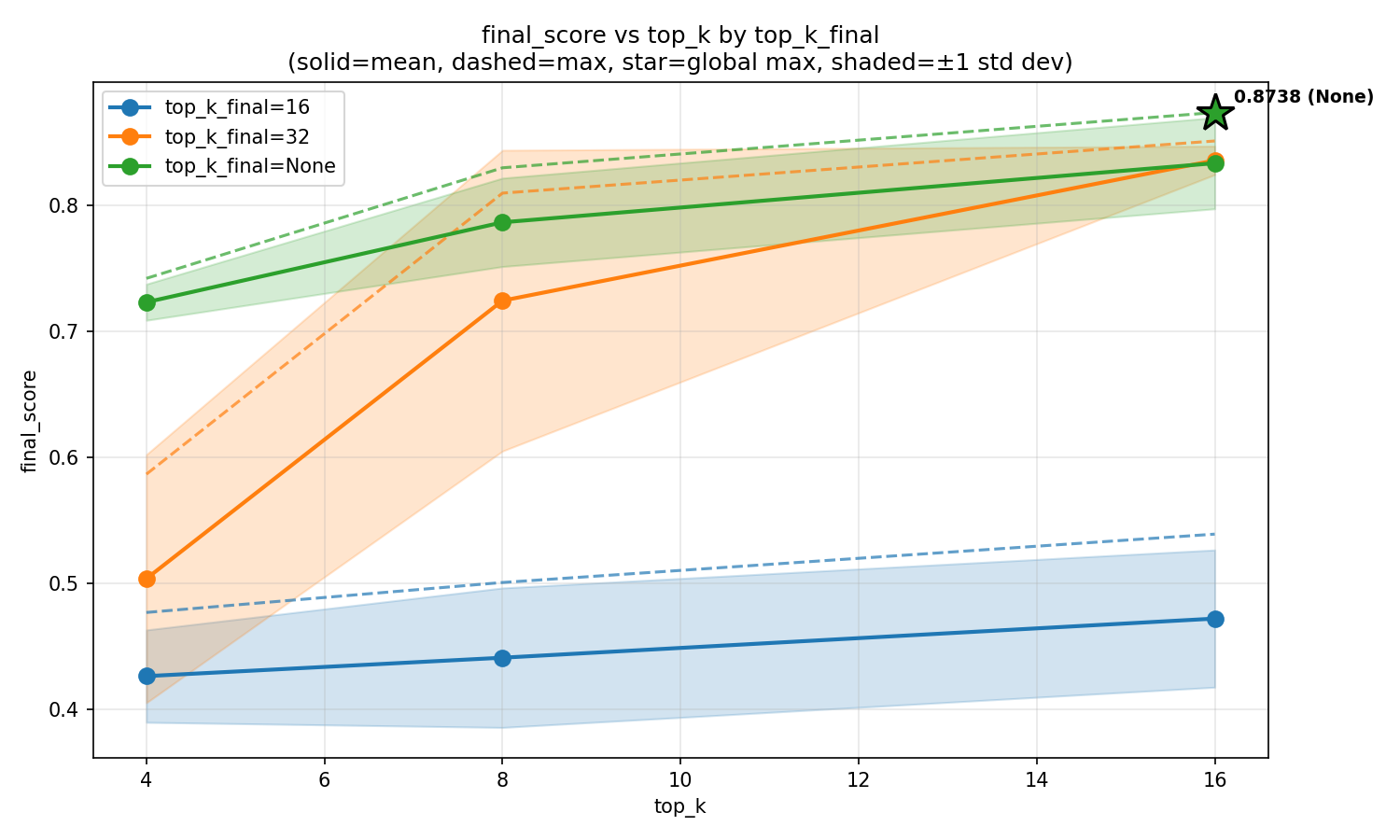}
\caption{Top-$k$ $\times$ Top-$k_{\text{final}}$.}
\label{fig:topk_final}
\end{minipage}
\end{figure}

Figures~\ref{fig:topk_retries}--\ref{fig:topk_final} show the impact of retry mechanism and final context truncation.

\paragraph{Retry Mechanism (Figure~\ref{fig:topk_retries}).} The retry mechanism provides substantial improvement, especially at low $k$ values (+69\% at $k$=4). When the model abstains due to insufficient evidence, expanding context and retrying often recovers the correct answer.

\paragraph{Final Top-$k$ (Figure~\ref{fig:topk_final}).} Truncating reranked results to top-$k_{\text{final}}$=32 provides optimal balance between coverage and precision. No truncation (using all retrieved results) slightly degrades performance due to noise accumulation.

\subsection{LLM Comparison}

\begin{figure}[H]
\centering
\includegraphics[width=0.7\textwidth]{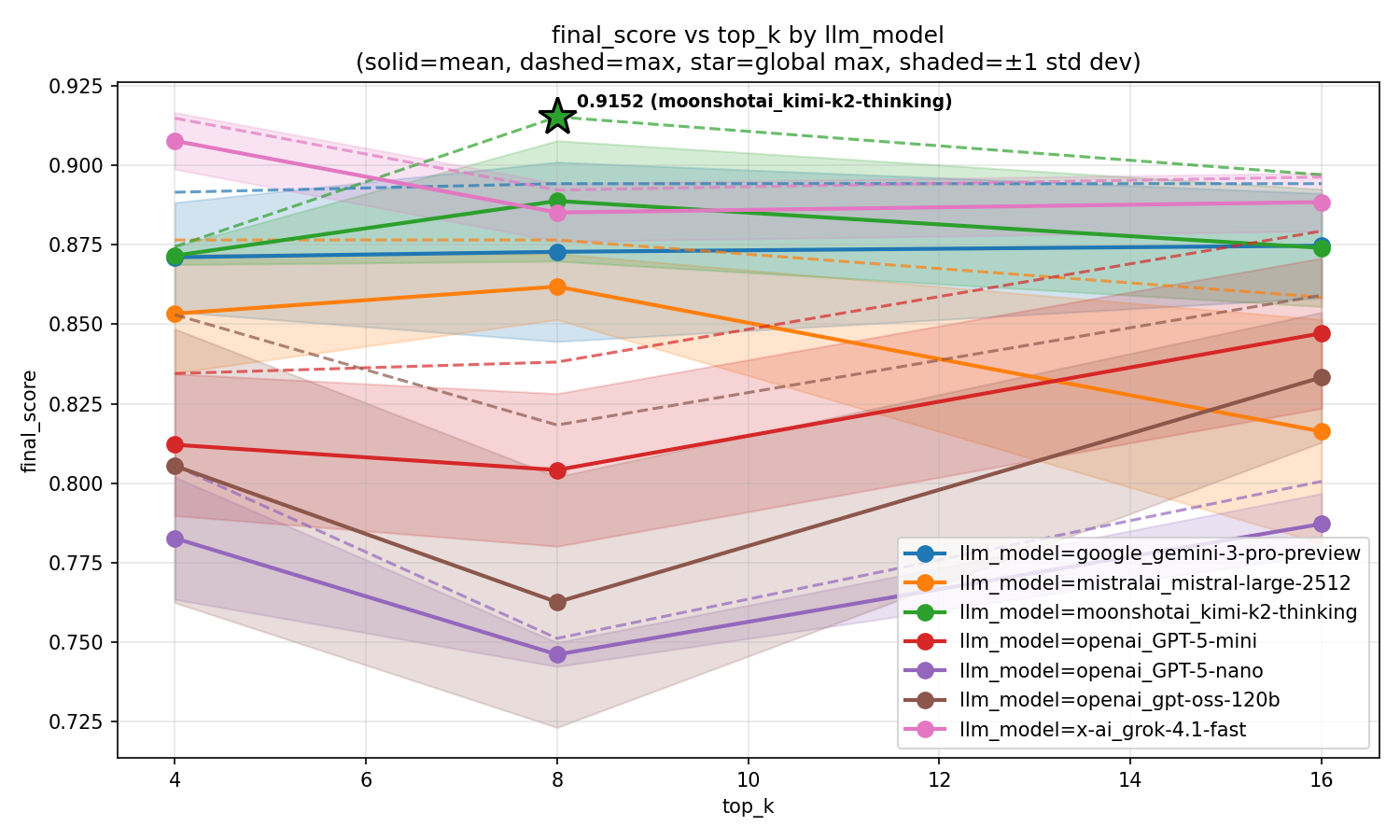}
\caption{LLM model $\times$ Top-$k$ configuration.}
\label{fig:llm_topk}
\end{figure}

Figure~\ref{fig:llm_topk} presents a comprehensive comparison across seven LLMs. Grok-4.1-fast achieves the highest overall performance, with Kimi-k2-thinking and Gemini-3-pro as close seconds. Notably, performance rankings vary with retrieval depth: Grok excels at $k$=4, while Gemini shows more consistent performance across all $k$ values.

\subsection{Ensemble Size Scaling}

\begin{figure}[H]
\centering
\begin{minipage}[t]{0.48\textwidth}
\centering
\includegraphics[width=\textwidth]{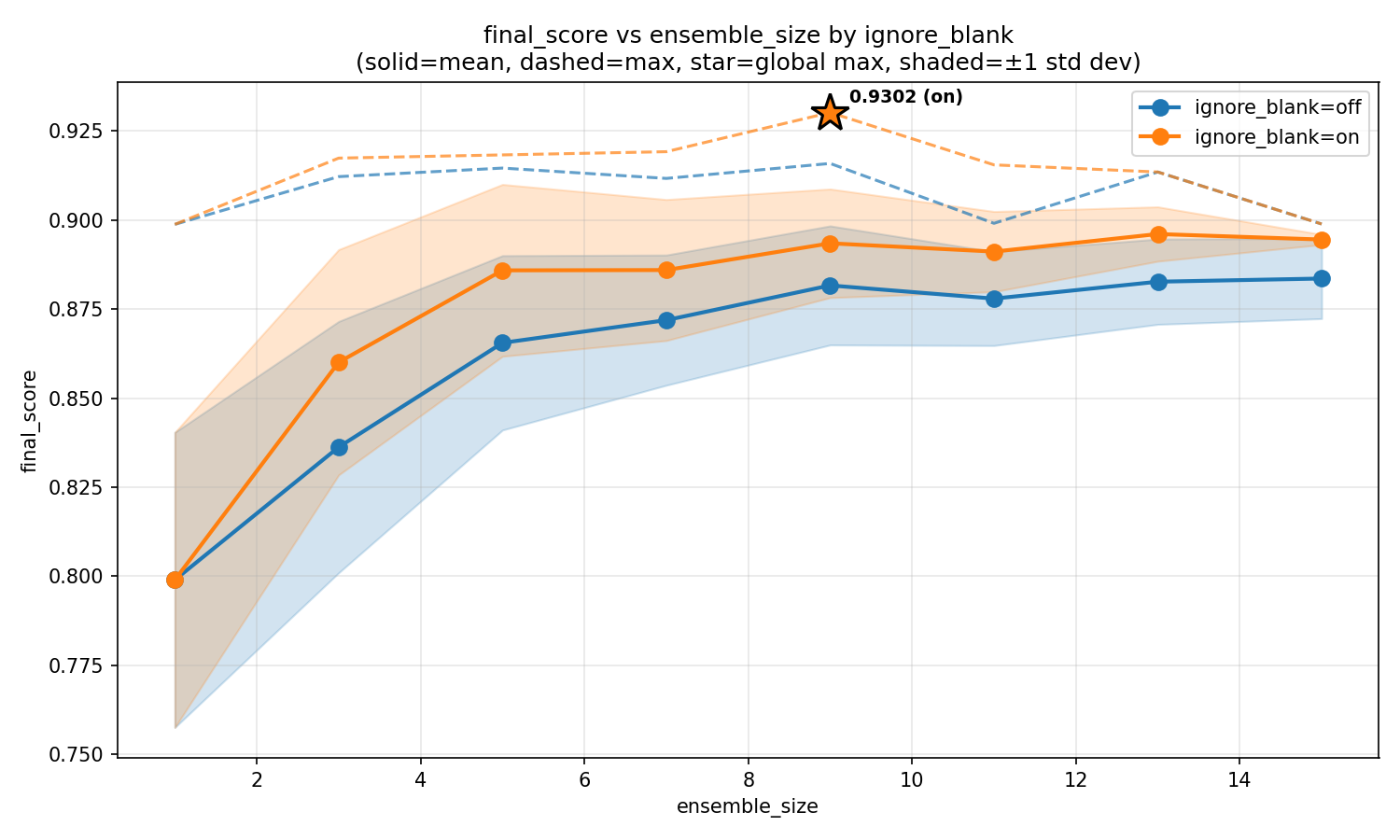}
\caption{Ensemble size $\times$ ignore\_blank.}
\label{fig:ensemble_ignore}
\end{minipage}
\hfill
\begin{minipage}[t]{0.48\textwidth}
\centering
\includegraphics[width=\textwidth]{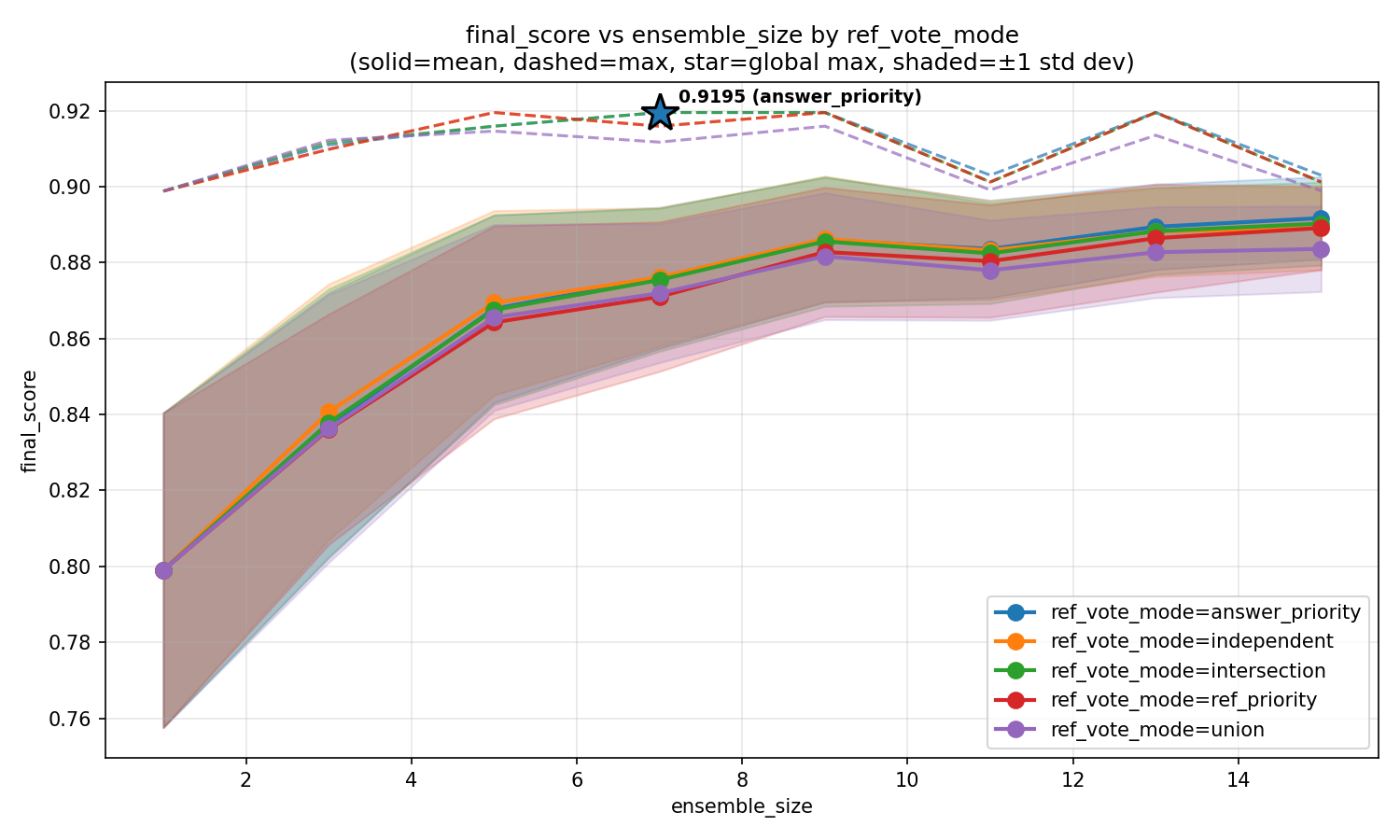}
\caption{Ensemble size $\times$ Voting strategy.}
\label{fig:ensemble_vote}
\end{minipage}
\end{figure}

Figures~\ref{fig:ensemble_ignore}--\ref{fig:ensemble_vote} and Tables~\ref{tab:ensemble_full}--\ref{tab:ignore_blank_full} present ensemble analysis.

\paragraph{Ensemble Scaling (Figure~\ref{fig:ensemble_ignore}).} Performance exhibits logarithmic scaling with ensemble size, with diminishing returns beyond $n$=9--11. The \texttt{ignore\_blank} option consistently improves performance by 1--2\%, with larger gains at smaller ensemble sizes where individual blank answers have more impact.

\paragraph{Voting Strategies (Figure~\ref{fig:ensemble_vote}).} \textsc{AnswerPriority} voting slightly outperforms other strategies, as it ensures reference consistency with the selected answer. \textsc{Union} voting underperforms due to including spurious references from minority votes.

\begin{table}[H]
\centering
\caption{Ensemble size vs. voting strategy (final score, mean over 32 subsamples).}
\label{tab:ensemble_full}
\small
\begin{tabular}{lcccccccc}
\toprule
\textbf{Strategy} & \textbf{$n$=1} & \textbf{$n$=3} & \textbf{$n$=5} & \textbf{$n$=7} & \textbf{$n$=9} & \textbf{$n$=11} & \textbf{$n$=13} & \textbf{$n$=15} \\
\midrule
Independent & .799 & .841 & .869 & .876 & .886 & .883 & .888 & .889 \\
RefPriority & .799 & .836 & .864 & .871 & .883 & .880 & .886 & .889 \\
AnswerPriority & .799 & .837 & .868 & .876 & .886 & .884 & \textbf{.889} & \textbf{.892} \\
Union & .799 & .836 & .866 & .872 & .882 & .878 & .883 & .884 \\
Intersection & .799 & .838 & .868 & .875 & .886 & .882 & .888 & .890 \\
\bottomrule
\end{tabular}
\end{table}

\begin{table}[H]
\centering
\caption{Effect of ignore\_blank on ensemble performance (union voting).}
\label{tab:ignore_blank_full}
\small
\begin{tabular}{lcccccccc}
\toprule
\textbf{ignore\_blank} & \textbf{$n$=1} & \textbf{$n$=3} & \textbf{$n$=5} & \textbf{$n$=7} & \textbf{$n$=9} & \textbf{$n$=11} & \textbf{$n$=13} & \textbf{$n$=15} \\
\midrule
Off & .799 & .836 & .866 & .872 & .882 & .878 & .883 & .884 \\
On & .799 & .860 & .886 & .886 & \textbf{.894} & .891 & .896 & .895 \\
$\Delta$ & -- & +.024 & +.020 & +.014 & +.012 & +.013 & +.013 & +.011 \\
\bottomrule
\end{tabular}
\end{table}

\section{Error Analysis}
\label{app:errors}

This appendix presents a comprehensive analysis of prediction errors across our hyperparameter sweeps,
providing insights into the failure modes of RAG systems and informing future improvements.
Understanding \textit{why} systems fail is as important as measuring \textit{how often} they succeed.

\subsection{Methodology and Error Categorization}

We analyzed 2,583 predictions across 63 sweep runs (7 LLMs $\times$ 3 top-$k$ settings $\times$ 3 random seeds)
to categorize failure modes. Each prediction was compared against ground truth using an automated
classification pipeline. Of all predictions, 75.2\% were fully correct, with the remaining 24.8\%
(641 errors) exhibiting various error types. Table~\ref{tab:error_summary} provides a high-level summary.

\begin{table}[ht]
\centering
\caption{Summary of error analysis across all sweep configurations.}
\label{tab:error_summary}
\small
\begin{tabular}{lc}
\toprule
\textbf{Metric} & \textbf{Value} \\
\midrule
Total predictions analyzed & 2,583 \\
Correct predictions & 1,942 (75.2\%) \\
Total errors & 641 (24.8\%) \\
\midrule
LLM configurations tested & 7 \\
Top-$k$ settings tested & 3 (4, 8, 16) \\
Random seeds per config & 3 \\
\bottomrule
\end{tabular}
\end{table}

\paragraph{Classification Pipeline.} We developed an automated error classification that assigns each incorrect prediction to exactly one category based on a deterministic decision tree:

\textit{Step 1 -- Abstention Check:} If ground truth is blank but prediction is not $\rightarrow$ \textbf{False positive}; if ground truth is not blank but prediction is blank $\rightarrow$ \textbf{False negative}; if both are blank $\rightarrow$ Correct.

\textit{Step 2 -- Value Comparison:} For non-blank predictions, we compute percentage error $\epsilon = |v_{\text{pred}} - v_{\text{true}}| / |v_{\text{true}}| \times 100\%$ and ratio $r = v_{\text{pred}} / v_{\text{true}}$ for numeric values; case-insensitive comparison for categorical values.

\textit{Step 3 -- Numeric Error Classification:} For $\epsilon > 0.1\%$ (WattBot tolerance): (1) \textbf{Unit conversion} if $r$ is within $\pm 5\%$ of any power of 10; (2) \textbf{Rounding/calculation} if $\epsilon \leq 10\%$; (3) \textbf{Value selection} otherwise.

\textit{Step 4 -- Reference and Type Errors:} If value is correct but references differ $\rightarrow$ \textbf{Reference mismatch}. If expected numeric but got text (or vice versa) $\rightarrow$ \textbf{Type mismatch}. If both non-numeric text but different $\rightarrow$ \textbf{Categorical mismatch}.

\subsection{Error Distribution and Patterns}

\begin{figure}[ht]
\centering
\includegraphics[width=0.75\textwidth]{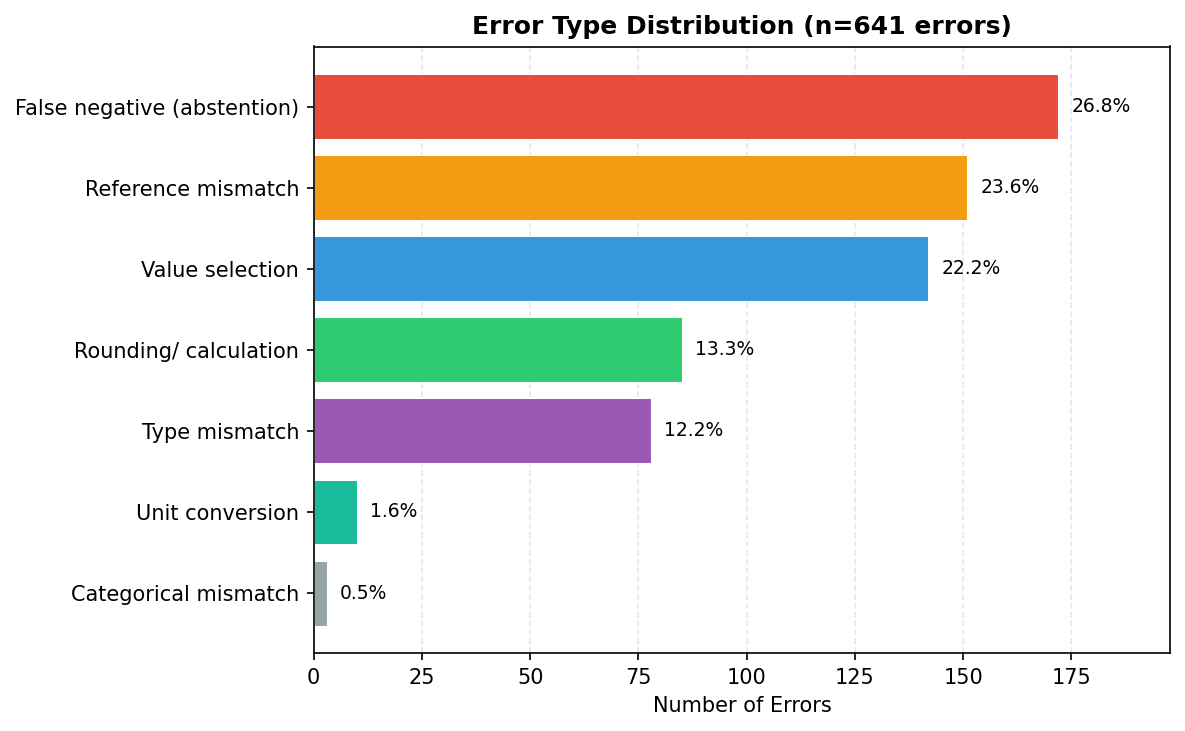}
\caption{Distribution of error types across 641 errors from 63 sweep runs. The top three categories
(abstention, reference mismatch, value selection) account for nearly three-quarters of all errors.}
\label{fig:error_dist}
\end{figure}

Applying this classification to all 641 errors yields the distribution shown in Figure~\ref{fig:error_dist}.
The three dominant categories, unnecessary abstention (26.8\%), reference mismatch (23.6\%), and
value selection (22.2\%), together account for 72.6\% of all errors, suggesting focused intervention
points for system improvement.

\begin{itemize}[noitemsep]
    \item \textbf{False negative (unnecessary abstention)}: 172 errors (26.8\%)
    \item \textbf{Reference ID mismatch}: 151 errors (23.6\%)
    \item \textbf{Value selection errors}: 142 errors (22.2\%)
    \item \textbf{Rounding/calculation errors}: 85 errors (13.3\%)
    \item \textbf{Type mismatch}: 78 errors (12.2\%)
    \item \textbf{Unit conversion errors}: 10 errors (1.6\%)
    \item \textbf{Categorical mismatch}: 3 errors (0.5\%)
\end{itemize}

\begin{figure}[ht]
\centering
\includegraphics[width=0.65\textwidth]{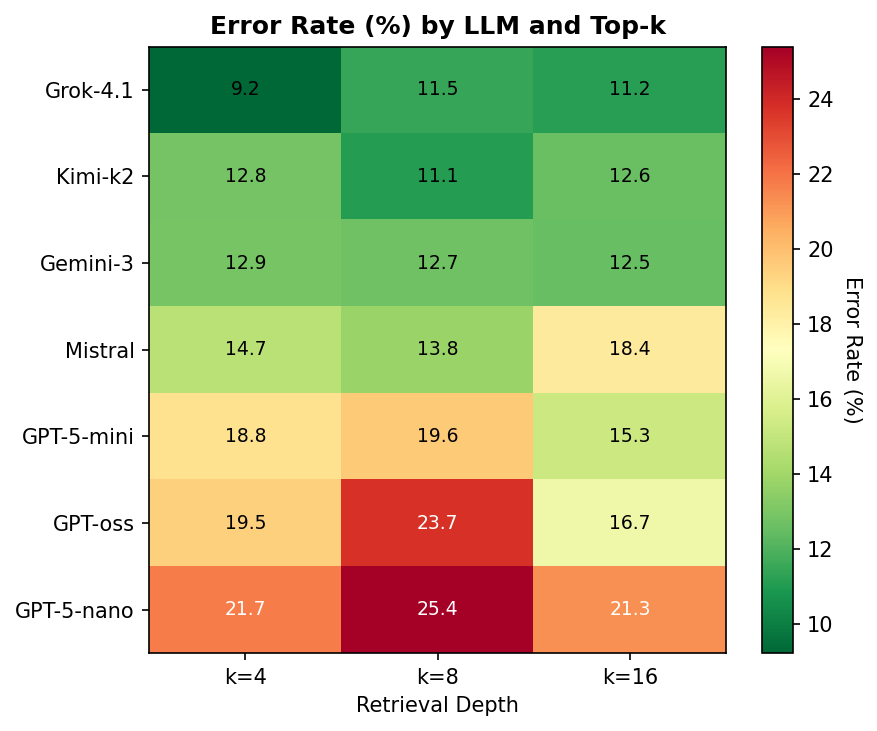}
\caption{Error rates by LLM model and retrieval depth. Grok-4.1-fast achieves the lowest error rates
across all configurations, while smaller models (GPT-5-nano) show consistently higher error rates.}
\label{fig:error_heatmap}
\end{figure}

\paragraph{Patterns by Configuration.} Beyond aggregate statistics, we examine how error rates vary across LLM models and retrieval depths.
Figure~\ref{fig:error_heatmap} shows the overall error rate (computed as $1 - \text{final\_score}$)
for each LLM and top-$k$ combination.

\begin{figure}[ht]
\centering
\begin{minipage}[t]{0.48\textwidth}
\centering
\includegraphics[width=\textwidth]{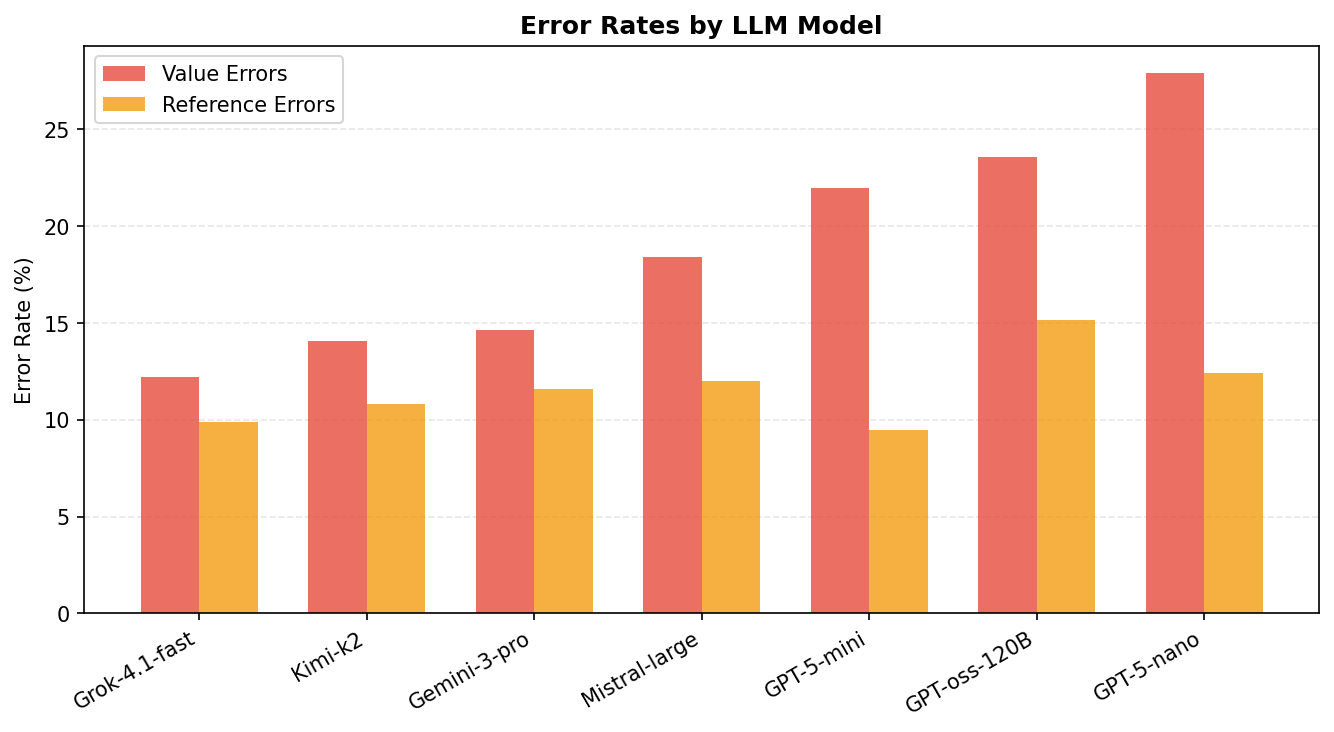}
\caption{Value vs.\ reference error rates by LLM. Value errors show larger variation across models.}
\label{fig:error_by_llm}
\end{minipage}
\hfill
\begin{minipage}[t]{0.48\textwidth}
\centering
\includegraphics[width=\textwidth]{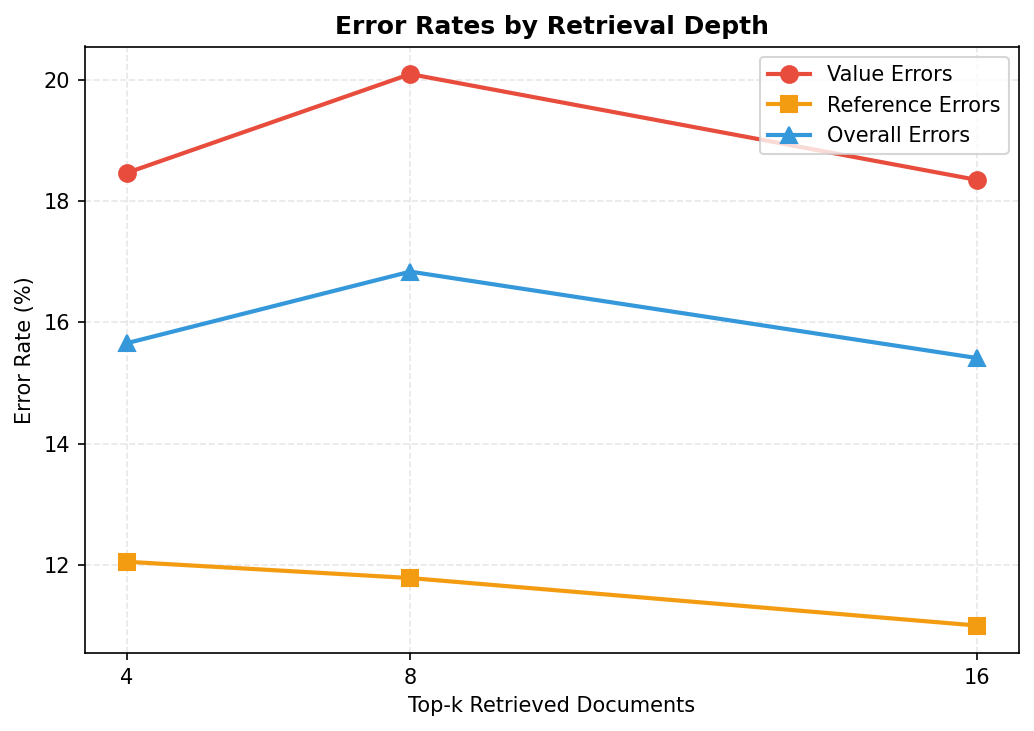}
\caption{Error rates by retrieval depth. $k=8$ shows highest value errors; reference errors decrease with $k$.}
\label{fig:error_by_topk}
\end{minipage}
\end{figure}

Figure~\ref{fig:error_by_llm} breaks down errors into value errors (wrong answer) and reference errors
(wrong citation) by model. Notably, value error rates vary more dramatically across models than reference
errors, suggesting that answer extraction capability is more model-dependent than citation accuracy.
Figure~\ref{fig:error_by_topk} shows how these error types vary with retrieval depth.

Several patterns emerge: (1) \textit{Model capacity matters}: larger models (Grok-4.1-fast, Kimi-k2, Gemini-3-pro)
achieve 9--13\% error rates compared to 21--25\% for GPT-5-nano. (2) \textit{Retrieval depth trade-offs}: the
relationship between top-$k$ and error rate is non-monotonic; $k=4$ and $k=16$ often outperform $k=8$.
(3) \textit{Error type differences}: value errors are more sensitive to model choice, while reference errors
are more stable across configurations.

\subsection{Concrete Examples and Analysis}

To illustrate our classification criteria, we provide representative examples from each error category:

\paragraph{False Negative (Unnecessary Abstention).}
\begin{quote}
\textbf{Question:} ``What is the carbon footprint (kgCO2eq) of training GPT-3?''\\
\textbf{Ground truth:} 552,000\\
\textbf{Prediction:} \texttt{is\_blank=true}\\
\textbf{Analysis:} The model abstained despite the answer appearing in the retrieved context. This often occurs when the exact phrasing differs (``carbon emissions'' vs ``carbon footprint'') or when the model is uncertain about unit conversions.
\end{quote}

\paragraph{Reference Mismatch.}
\begin{quote}
\textbf{Question:} ``What is the PUE of Google's data centers?''\\
\textbf{Ground truth value:} 1.1, \textbf{refs:} [google2023]\\
\textbf{Prediction value:} 1.1, \textbf{refs:} [patterson2021]\\
\textbf{Analysis:} The answer is correct, but the model cited a survey paper that \textit{mentions} Google's PUE rather than Google's official report. Both documents were in the retrieved context.
\end{quote}

\paragraph{Value Selection Error.}
\begin{quote}
\textbf{Question:} ``What is the energy consumption (kWh) per inference for GPT-4?''\\
\textbf{Ground truth:} 0.0029\\
\textbf{Prediction:} 0.0017 ($\epsilon = 41.4\%$, $r = 0.59$)\\
\textbf{Analysis:} The retrieved context contained energy figures for multiple model sizes. The model selected the value for a smaller variant rather than the full GPT-4 model. Since $r = 0.59$ is not close to any power of 10, this is classified as value selection rather than unit conversion.
\end{quote}

\paragraph{Rounding/Calculation Error.}
\begin{quote}
\textbf{Question:} ``What percentage of total US electricity consumption is attributed to data centers?''\\
\textbf{Ground truth:} 2.5\\
\textbf{Prediction:} 2.7 ($\epsilon = 8\%$)\\
\textbf{Analysis:} The model retrieved the correct source but performed a slightly incorrect calculation or read from an outdated figure in the same document. Since $\epsilon \leq 10\%$, this is classified as rounding/calculation error.
\end{quote}

\paragraph{Unit Conversion Error.}
\begin{quote}
\textbf{Question:} ``How much water (liters) does training a large language model consume?''\\
\textbf{Ground truth:} 5,439,000\\
\textbf{Prediction:} 5,439 ($r = 0.001 = 10^{-3}$)\\
\textbf{Analysis:} The model confused ``million liters'' with ``liters.'' Since the ratio is within $\pm 5\%$ of $10^{-3}$, this is classified as unit conversion error.
\end{quote}

\paragraph{Type Mismatch.}
\begin{quote}
\textbf{Question:} ``What is the typical GPU utilization rate in data centers?''\\
\textbf{Ground truth:} 30 (numeric percentage)\\
\textbf{Prediction:} ``30-50\%'' (range string)\\
\textbf{Analysis:} The ground truth expects a single numeric value, but the model output a range. This is classified as type mismatch regardless of whether the correct value falls within the range.
\end{quote}

\paragraph{Discussion.} Each dominant error category suggests different intervention strategies:
\textbf{Unnecessary abstention} (26.8\%) occurs when the model outputs \texttt{is\_blank} despite sufficient context; the retry mechanism (Section~\ref{subsec:context}) directly addresses this by expanding context when the model abstains.
\textbf{Reference mismatch} (23.6\%) indicates provenance tracking remains challenging; common causes include citing survey papers rather than original sources.
\textbf{Value selection errors} (22.2\%) arise when context contains multiple plausible values and the model selects the wrong one.
\textbf{Rounding/calculation errors} (13.3\%) are close misses indicating the system found correct information but made minor arithmetic mistakes.
\textbf{Unit conversion errors} are notably rare (1.6\%), indicating LLMs handle unit consistency well.

\subsection{Implications for RAG System Design}

Our error analysis suggests several directions for improving RAG system reliability:

\begin{enumerate}[noitemsep]
\item \textbf{Reduce unnecessary abstention.} The retry mechanism with context expansion proves effective; further improvements could include confidence calibration or ensemble voting that ignores blank answers.

\item \textbf{Improve reference tracking.} Reference mismatch errors suggest LLMs struggle with precise attribution. Potential solutions include explicit reference extraction as a separate step or constraining answers to specific retrieved documents.

\item \textbf{Disambiguate similar values.} Value selection errors often occur when context contains multiple related values. Query planning with more specific sub-queries could help.

\item \textbf{Accept minor calculation errors.} Rounding errors represent fundamentally correct retrievals; evaluation metrics could distinguish between major and minor errors.

\item \textbf{Model selection matters.} The 16-point gap between Grok-4.1-fast (9.2\%) and GPT-5-nano (25.4\%) underscores that LLM capability is a dominant factor, potentially more impactful than retrieval optimizations.
\end{enumerate}

\section{WattBot 2025 Leaderboard Analysis}
\label{app:leaderboard}

This appendix presents a comprehensive analysis of the WattBot 2025 competition leaderboard, 
examining the dynamics between public and private evaluation partitions. Understanding these 
dynamics is crucial for interpreting competition results and provides insights into method 
robustness, a key concern when deploying RAG systems in production environments where 
evaluation data may differ from development conditions.

\subsection{Final Leaderboard}
\begin{figure}[H]
    \centering
    \begin{subfigure}[t]{0.48\textwidth}
        \centering
        \includegraphics[width=\textwidth]{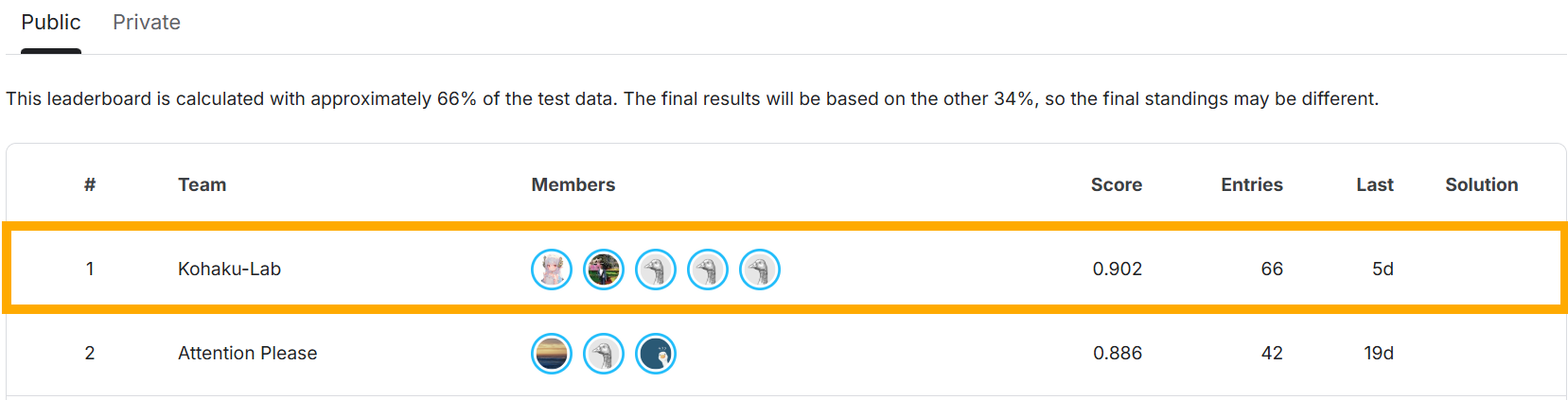}
        \caption{Public leaderboard}
        \label{fig:leaderboard-public}
    \end{subfigure}
    \hfill
    \begin{subfigure}[t]{0.48\textwidth}
        \centering
        \includegraphics[width=\textwidth]{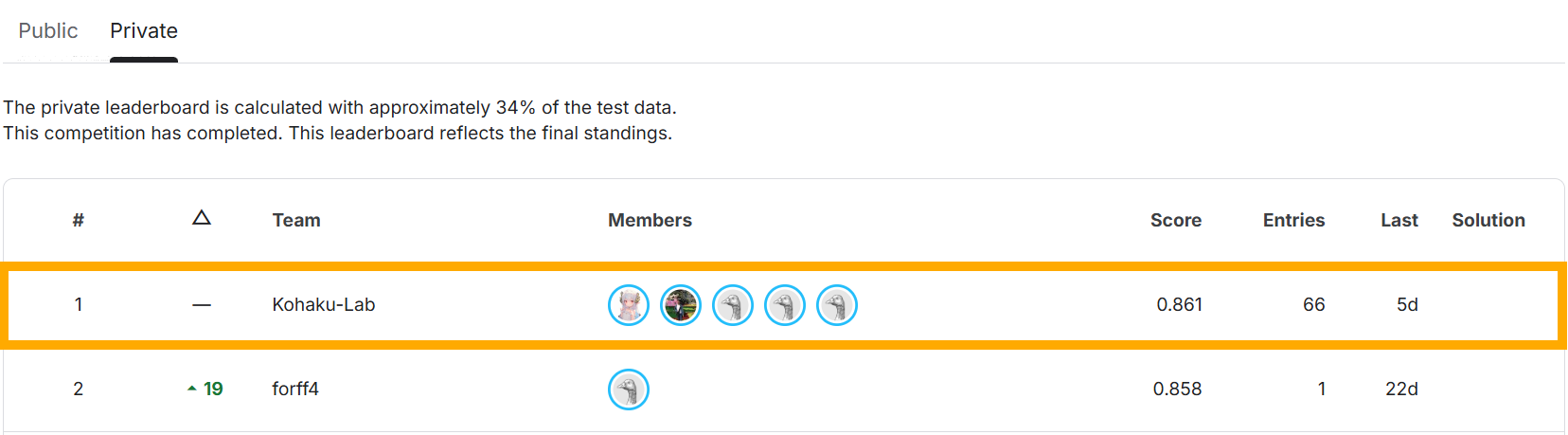}
        \caption{Private leaderboard (final)}
        \label{fig:leaderboard-private}
    \end{subfigure}
    \caption{WattBot 2025 Challenge leaderboards. (a) Public leaderboard visible during the competition. (b) Final private leaderboard revealed after the submission deadline. KohakuRAG maintained rank \#1 across both partitions.}
    \label{fig:leaderboard}
\end{figure}

Figure~\ref{fig:leaderboard} shows the official leaderboards of the WattBot 2025 Challenge. KohakuRAG (team ``Kohaku-Lab'') achieved first place on both the public and private leaderboards, with scores of 0.902 and 0.861 respectively.

\subsection{Analysis Methodology}

The WattBot 2025 competition employed a standard Kaggle-style evaluation protocol with two
distinct test partitions: a \textit{public} partition (66\% of test questions) whose scores were
visible during the competition, and a \textit{private} partition (34\%) revealed only after the
submission deadline. Importantly, \textbf{both partitions are held-out test sets}: neither is
directly accessible for training or validation. The key difference is that teams can observe
public scores and iterate their submissions accordingly, while private scores remain completely
hidden until the competition ends.

This creates an interesting evaluation dynamic: the public leaderboard measures performance on
one test distribution that teams can \textit{indirectly} optimize through submission feedback,
while the private leaderboard measures transfer to a different test distribution with no
optimization signal. A robust method should perform well on \textbf{both} partitions, as the
final evaluation considers aggregate performance across the entire test set.

We analyze the complete leaderboard data to understand:
\begin{itemize}[noitemsep]
    \item How consistently do methods perform across the two test partitions?
    \item What patterns distinguish teams with stable vs.\ volatile performance?
    \item How does KohakuRAG's performance compare across both evaluation dimensions?
\end{itemize}

\subsection{Score Transition Analysis}

The competition included 66 teams that submitted predictions evaluated on both
partitions. Table~\ref{tab:leaderboard_stats} summarizes the key statistics describing
score transitions from public to private evaluation.

\begin{table}[h]
\centering
\caption{Summary statistics for public-to-private score transitions across all participating teams.
The score delta ($\Delta$) is computed as private score minus public score, where negative values
indicate performance degradation on the private partition.}
\label{tab:leaderboard_stats}
\small
\begin{tabular}{lc}
\toprule
\textbf{Metric} & \textbf{Value} \\
\midrule
Total teams & 66 \\
Mean $\Delta$ (private $-$ public) & -0.0106 \\
Standard deviation of $\Delta$ & 0.0452 \\
Median $\Delta$ & -0.0160 \\
Minimum $\Delta$ & -0.2130 \\
Maximum $\Delta$ & 0.0980 \\
\midrule
Teams with score decrease & 43 (65.2\%) \\
Teams with score increase & 22 (33.3\%) \\
Pearson correlation (public score, $\Delta$) & -0.389 \\
\bottomrule
\end{tabular}
\end{table}

Several observations emerge from these statistics. First, the majority of teams
(65\%) experienced score degradation on the private partition, with a mean delta of $-0.0106$.
This systematic decrease is expected: teams naturally gravitate toward configurations that
score well on the observable public partition, which may not transfer perfectly to the
private partition's question distribution.

\begin{figure}[h]
\centering
\includegraphics[width=0.7\textwidth]{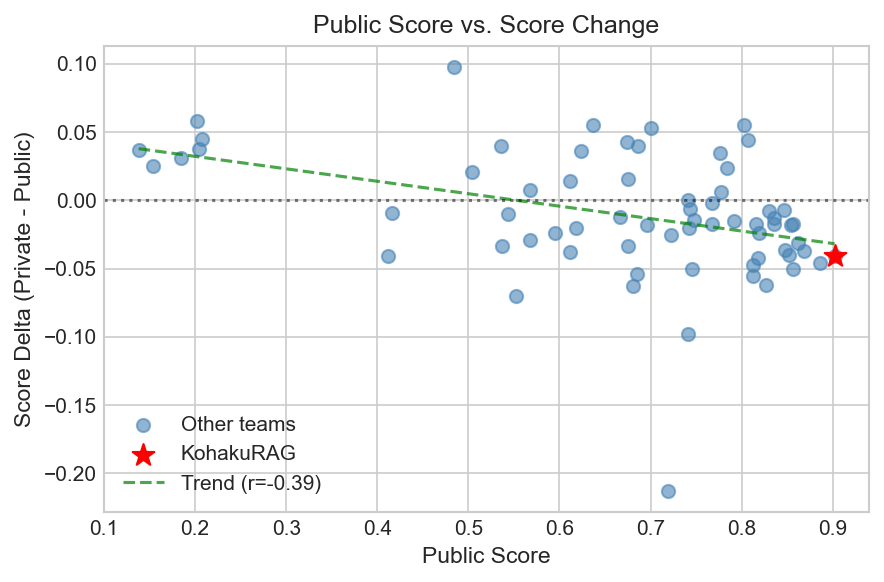}
\caption{Relationship between public score and score delta. Each point represents a team;
KohakuRAG is marked with a red star. The negative trend reflects variance between the two
test distributions rather than indicating that high public scores are inherently problematic.}
\label{fig:lb_correlation}
\end{figure}

Second, we observe a moderate negative correlation ($r = -0.389$) between public scores and
score deltas. This indicates that teams with higher public scores tended to experience larger
drops on the private partition. However, this does \textit{not} imply that achieving high
public scores is undesirable; it simply reflects that there is variance between the two
test distributions, and methods optimized (even indirectly) for one may not perfectly transfer
to the other. Figure~\ref{fig:lb_correlation} visualizes this relationship.

\begin{figure}[h]
\centering
\includegraphics[width=0.7\textwidth]{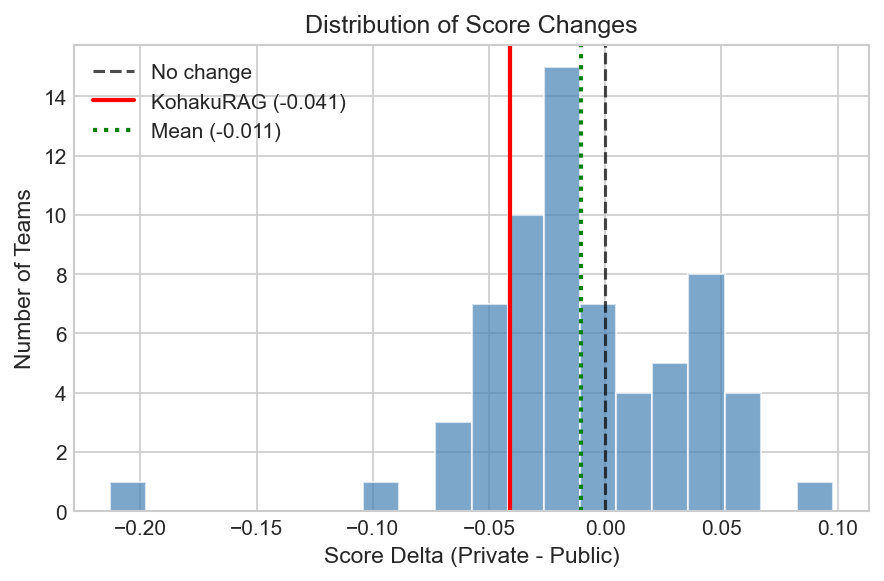}
\caption{Distribution of score deltas (private $-$ public) across all teams.
KohakuRAG's delta is marked for reference.}
\label{fig:lb_delta_hist}
\end{figure}

Figure~\ref{fig:lb_delta_hist} shows the distribution of score deltas across all teams,
providing context for evaluating any individual team's performance transition. The distribution
is left-skewed, with more teams experiencing score drops than gains.

\subsection{Rank Movement Patterns}

\paragraph{Top-10 Teams.} To better understand the dynamics among competitive submissions, we examine the top 10 teams
from the public leaderboard and their corresponding private performance. Table~\ref{tab:top10_transition}
presents these transitions in detail.

\begin{table}[h]
\centering
\caption{Public-to-private transition for the top 10 public leaderboard teams. The $\Delta$ column
shows score change, and Rank $\Delta$ shows position change (positive = improved). KohakuRAG ($\star$)
was the only team to maintain first place across both partitions.}
\label{tab:top10_transition}
\small
\begin{tabular}{cccccc}
\toprule
\textbf{Pub Rank} & \textbf{Pub Score} & \textbf{Priv Score} & \textbf{Priv Rank} & \textbf{Score $\Delta$} & \textbf{Rank $\Delta$} \\
\midrule
\#1 $\star$ & 0.902 & 0.861 & \#1 & -0.041 & 0 \\
\#2 & 0.886 & 0.840 & \#4 & -0.046 & -2 \\
\#3 & 0.868 & 0.831 & \#9 & -0.037 & -6 \\
\#4 & 0.862 & 0.831 & \#8 & -0.031 & -4 \\
\#5 & 0.856 & 0.806 & \#17 & -0.050 & -12 \\
\#6 & 0.856 & 0.839 & \#6 & -0.017 & 0 \\
\#7 & 0.854 & 0.836 & \#7 & -0.018 & 0 \\
\#8 & 0.852 & 0.812 & \#13 & -0.040 & -5 \\
\#9 & 0.848 & 0.812 & \#14 & -0.036 & -5 \\
\#10 & 0.846 & 0.839 & \#5 & -0.007 & +5 \\
\bottomrule
\end{tabular}
\end{table}

The top 10 public teams collectively show a mean delta of $-0.0323$, larger in magnitude
than the overall mean of $-0.0106$. This is consistent with the correlation analysis: teams
at the performance frontier tend to show more variance between partitions. Importantly, despite
these deltas, the top public teams generally remained competitive on the private leaderboard
as well; the correlation reflects \textit{relative} shifts, not absolute failure.

\begin{figure}[h]
\centering
\includegraphics[width=0.75\textwidth]{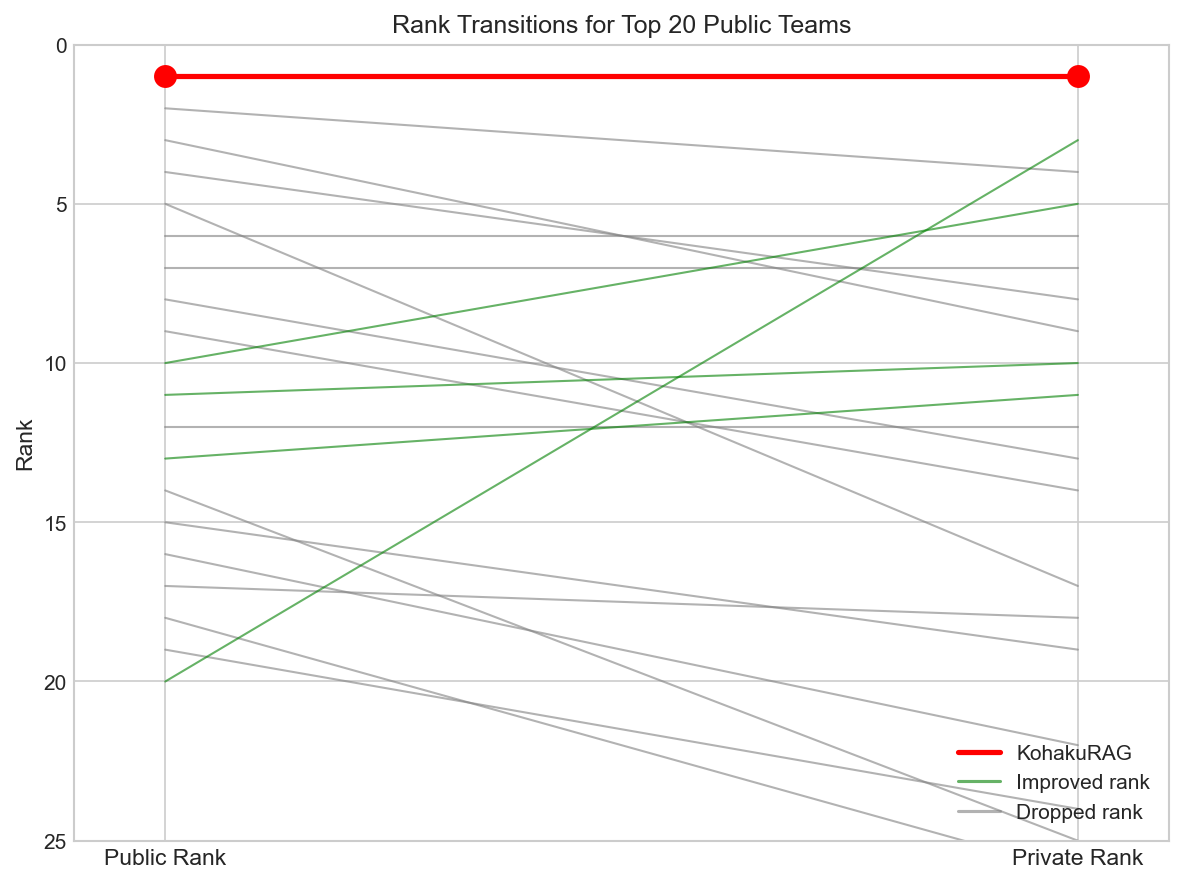}
\caption{Rank transitions from public to private leaderboard for top 20 public teams.
Lines connect each team's public rank (left) to private rank (right). KohakuRAG (red)
maintained rank \#1; green lines indicate teams that improved, gray lines indicate teams that dropped.}
\label{fig:lb_rank_flow}
\end{figure}

Figure~\ref{fig:lb_rank_flow} visualizes rank transitions for the top 20 public teams,
illustrating the substantial reshuffling that occurred when private scores were revealed.

\paragraph{Stability by Rank Band.} To determine whether the pattern is concentrated among top performers or distributed
throughout the leaderboard, we partition teams into rank bands (Table~\ref{tab:delta_by_band}).

\begin{table}[h]
\centering
\caption{Score delta statistics stratified by public rank band. Top-ranked teams show more negative
mean deltas, while mid-to-low ranked teams show near-zero or positive mean deltas.}
\label{tab:delta_by_band}
\small
\begin{tabular}{lcccc}
\toprule
\textbf{Rank Band} & \textbf{$n$} & \textbf{Mean $\Delta$} & \textbf{Std $\Delta$} & \textbf{Median $\Delta$} \\
\midrule
1--10 & 10 & -0.0323 & 0.0132 & -0.0365 \\
11--20 & 10 & -0.0241 & 0.0289 & -0.0205 \\
21--30 & 10 & +0.0016 & 0.0284 & -0.0040 \\
31--50 & 20 & -0.0181 & 0.0597 & -0.0190 \\
51--100 & 16 & +0.0131 & 0.0414 & +0.0230 \\
\bottomrule
\end{tabular}
\end{table}

Teams in the top rank bands (1--10 and 11--20) exhibit more negative mean deltas compared to
lower-ranked teams. The 21--30 and 51--100 bands show near-zero or slightly positive mean deltas.
This pattern has a natural explanation: lower-ranked teams had more ``room to improve'' on the
private partition (regression to the mean), while top teams were already near the performance
ceiling and thus more likely to show decreases. This does \textit{not} suggest that targeting
mid-tier public rankings leads to better outcomes, since teams that jumped in rank still achieved
lower final scores than consistently top-performing teams like KohakuRAG.

\paragraph{Largest Rank Changes.} Table~\ref{tab:rank_movement} highlights the most dramatic rank changes in both directions.

\begin{table}[h]
\centering
\caption{Teams with the largest rank improvements (left) and largest rank drops (right).
Notable gainers came from outside the top 20 public, while notable losers fell from top-10 positions.}
\label{tab:rank_movement}
\small
\begin{tabular}{cccc}
\toprule
\textbf{Public} & \textbf{Private} & \textbf{$\Delta$ Rank} & \textbf{$\Delta$ Score} \\
\midrule
\#21 & \#2 & +19 & +0.055 \\
\#20 & \#3 & +17 & +0.044 \\
\#25 & \#15 & +10 & +0.035 \\
\#23 & \#16 & +7 & +0.024 \\
\#10 & \#5 & +5 & -0.007 \\
\bottomrule
\end{tabular}
\hfill
\begin{tabular}{cccc}
\toprule
\textbf{Public} & \textbf{Private} & \textbf{$\Delta$ Rank} & \textbf{$\Delta$ Score} \\
\midrule
\#5 & \#17 & $-$12 & -0.050 \\
\#14 & \#25 & $-$11 & -0.062 \\
\#18 & \#26 & $-$8 & -0.055 \\
\#3 & \#9 & $-$6 & -0.037 \\
\#16 & \#22 & $-$6 & -0.042 \\
\bottomrule
\end{tabular}
\end{table}

The most dramatic rank improvements came from teams initially ranked \#20--25, with the
largest gainer jumping 19 positions from \#21 to \#2. However, it is important to note
that these teams' \textit{final private scores} (0.858, 0.851) were still lower than
KohakuRAG's (0.861). The rank improvement reflects relative movement, not absolute
superiority; their methods happened to transfer better to the private partition's
specific question distribution, but they did not achieve the highest overall performance.

Conversely, some top-10 public teams experienced significant drops, with one team falling
12 positions. These cases illustrate the variance between test partitions, but do not
indicate that high public performance is undesirable. The goal remains to maximize
performance on \textbf{both} partitions, which KohakuRAG achieved by maintaining \#1
ranking across both.

\subsection{KohakuRAG Performance and Implications}

\paragraph{Our Results.} KohakuRAG achieved a public score of 0.902 (rank \#1) and a private score of 0.861 (rank \#1),
with a delta of $-0.041$. The key result is that \textbf{KohakuRAG achieved the highest score
on both partitions independently}, maintaining first place across both test distributions.
This is the ideal outcome: strong performance that transfers consistently.

While our delta magnitude is larger than 68\% of teams, this metric is less meaningful than
absolute performance. Many teams with small deltas simply had mediocre scores on both
partitions. What matters is achieving high scores on \textit{both} test sets, which KohakuRAG
accomplished.

Our ensemble-based approach contributed to this consistency. By aggregating predictions from
multiple model runs with abstention-aware voting, we reduced variance in individual predictions.
The ensemble does not guarantee zero delta, as the two test partitions genuinely differ in their
question distributions, but it helps ensure that performance remains competitive across
both evaluation conditions rather than being optimized for one at the expense of the other.

\paragraph{Broader Implications.} The patterns observed have implications for RAG system development:
\begin{enumerate}[noitemsep]
\item \textbf{Both test partitions matter equally.} Teams that ``improved'' from public to private
often did so from a lower baseline and still achieved lower final scores than consistently top-performing teams.

\item \textbf{Rank changes reflect distribution variance, not method quality.} The teams that jumped from
\#20$\rightarrow$\#2 did not develop superior approaches; they happened to have methods
that transferred better to the private partition's specific characteristics.

\item \textbf{Ensemble methods help maintain consistency.} KohakuRAG's ensemble approach
contributed to achieving \#1 on both partitions by reducing variance across evaluation conditions.

\item \textbf{Absolute performance trumps relative stability.} Having a small delta is
meaningless if both scores are low. The objective is maximizing aggregate performance on both partitions.
\end{enumerate}

\section{Limitations}
\label{sec:limitations}

Our work has several limitations worth noting. Our document parsing pipeline relies on rule-based heuristics for section detection, paragraph segmentation, and sentence boundary identification. While effective for well-structured technical documents, these heuristics may fail on irregular layouts, multi-column formats, or documents with non-standard visual hierarchies, propagating errors to downstream retrieval.

Reference attribution remains challenging. Our error analysis reveals that 23.6\% of errors involve citation mismatch—the LLM produces correct answers but cites survey papers or secondary sources rather than original documents. This occurs because multiple documents in the corpus discuss overlapping topics, and the model lacks explicit mechanisms to prefer primary sources over derivative ones.

Ensemble inference, while improving robustness, incurs significant computational overhead. Our best configurations require 7--15 independent runs, linearly scaling both latency and API costs. For latency-sensitive or cost-constrained deployments, this overhead may be prohibitive despite the accuracy gains.

Our multimodal integration is relatively shallow. Image captions are generated independently without document context, and direct image retrieval yielded only marginal improvements (+1.2\%) over caption-based approaches. Complex visualizations such as multi-panel figures or data-dense charts remain challenging to retrieve and interpret effectively.

Finally, our evaluation is confined to a single domain—AI energy consumption across 32 documents. While we achieve strong results on WattBot 2025, generalization to other domains, larger corpora, or different question types remains unvalidated.

\section{Future Work}
\label{sec:future_work}

Several directions merit further investigation. Replacing rule-based parsing with learned approaches such as LayoutLM or DocFormer could improve robustness to diverse document layouts and reduce error propagation from incorrect structure extraction.

Addressing reference mismatch may require reference-constrained decoding that restricts citations to top-ranked passages, or explicit attribution verification as a post-processing step. Training with citation supervision could also improve the model's ability to identify primary sources.

The computational cost of ensemble inference motivates research into efficient alternatives. Confidence-based selective ensembling could apply majority voting only for uncertain predictions, while knowledge distillation could transfer ensemble robustness into a single model. Adaptive retrieval that predicts when additional context is needed—rather than using fixed $k$—could further reduce unnecessary computation.

Deeper multimodal integration through cross-modal attention and structured extraction of charts and tables could better capture visual information in technical documents. Finally, validation on diverse benchmarks spanning multiple domains, scales, and question types would establish broader applicability of our approach.

\end{document}